\documentclass[acmtog]{acmart} %

\usepackage[normalem]{ulem}

\usepackage{booktabs} %
\usepackage{float}
\usepackage{color,colortbl}
\usepackage{multirow}
\usepackage{enumitem}
\usepackage{graphicx}
\usepackage[normalem]{ulem}
\useunder{\uline}{\ul}{}
\usepackage[table,xcdraw]{xcolor}
\usepackage[export]{adjustbox}
\usepackage{subcaption}
\usepackage[ruled]{algorithm2e} %
\usepackage[capitalize]{cleveref}

\usepackage{pifont}    %
\newcommand{\cmark}{\textcolor{green!60!black}{\ding{51}}}  %
\newcommand{\xmark}{\textcolor{red!70!black}{\ding{55}}}    %

\SetAlFnt{\small}
\SetAlCapFnt{\small}
\SetAlCapNameFnt{\small}
\SetAlCapHSkip{0pt}
\theoremstyle{definition}

\acmJournal{TOG}
\acmYear{2025}

\setcopyright{none}

\acmDOI{0000001.0000001}

\definecolor{PositiveColor}{rgb}{.7,1,.7}
\definecolor{NegativeColor}{rgb}{1,.7,.7}
\definecolor{darkblue}{rgb}{0.2,0.4,0.6}
\definecolor{darkgreen}{rgb}{0.2,0.5,0.2}

\newcommand{\review}[1]{{\color{blue}#1}\normalfont}
\renewcommand{\review}[1]{#1}{}

\newcommand{\model}{WiP }
\newcommand{\modelNoSpace}{WiP}
\begin{document}

\title[Mocap \review{Anywhere: Towards} Pairwise-Distance based Motion Capture in the Wild (\emph{for the Wild})]{Mocap \review{Anywhere: Towards} Pairwise-Distance based Motion Capture \\ in the Wild (\emph{for the Wild})}

\author{Ofir Abramovich}
\orcid{0009-0007-4248-6423}
\affiliation{%
  \institution{Reichman University}
  \city{Herzliya}
  \postcode{4610101}
  \country{Israel}}
\affiliation{%
  \institution{CYENS Centre of Excellence}
  \city{Nicosia}
  \postcode{1016}
  \country{Cyprus}}
\email{ofir1080@gmail.com}

\author{Ariel Shamir}
\orcid{000-0001-7082-7845}
\affiliation{%
  \institution{Reichman University}
  \city{Herzliya}
  \postcode{4610101}
  \country{Israel}}
\email{arik@runi.ac.il}

\author{Andreas Aristidou}
\orcid{0000-0001-7754-0791}
\affiliation{
  \institution{University of Cyprus}
  \city{Nicosia}
  \postcode{1678}
  \country{Cyprus}
}
\affiliation{%
  \institution{CYENS Centre of Excellence}
  \city{Nicosia}
  \postcode{1016}
  \country{Cyprus}
}
\email{a.aristidou@ieee.org}

\renewcommand\shortauthors{Abramovich et al.}

\begin{abstract}
We introduce a novel motion capture system that reconstructs full-body 3D motion using only sparse pairwise distance (PWD) measurements from body-mounted %
sensors. Using time-of-flight ranging between wireless nodes, our method eliminates the need for external cameras, enabling robust operation in uncontrolled and outdoor environments. Unlike traditional optical or inertial systems, our approach is shape-invariant and resilient to environmental constraints such as lighting and magnetic interference. At the core of our system is Wild-Poser (\model for short), a compact, real-time Transformer-based architecture that directly predicts 3D joint positions from noisy or corrupted PWD measurements, \review{which can later be used for joint rotation reconstruction via learned methods}. \model~generalizes across subjects of varying morphologies, including non-human species, without requiring individual body measurements or shape fitting. Operating in real time, \model achieves low joint position error and demonstrates accurate 3D motion reconstruction for both human and animal subjects in-the-wild. Our empirical analysis highlights its potential for scalable, low-cost, and general-purpose motion capture in real-world settings.
\end{abstract}



\begin{teaserfigure}
    \centering
    \includegraphics[width=\textwidth]{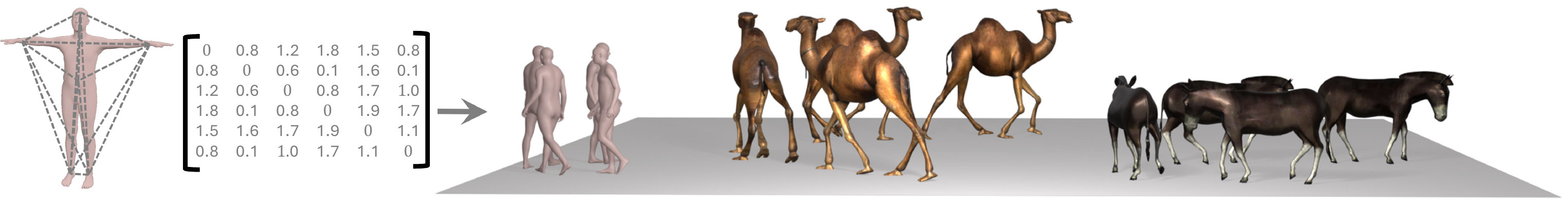}
    \caption{We present a motion capture method based on inter-joint pairwise distance measurements over time. Our method allows reconstructing pose and motion of different articulated subjects, in the wild.}
    \label{fig:teaser}
\end{teaserfigure}

\maketitle

\section{Introduction}
Motion capture (mocap) is the primary technology for animation and character control, biomechanical analysis, and immersive experiences. It transforms the movement of a subject into 3D positional and/or rotational data, enabling highly realistic representations of motion. Although widely adopted in fields such as film, games, sports science, and virtual production, mocap systems remain constrained by cost, scalability, and environmental requirements.

Today, the most precise systems in use are optical systems that rely on multiple cameras to triangulate markers worn by the subject. While these systems (e.g., Vicon, OptiTrack, PhaseSpace) offer millimeter-level accuracy, they are expensive, require multi-camera calibration, and are typically limited to indoor, studio-controlled environments with fixed lighting conditions. To address portability and cost limitations, IMU-based systems (e.g., Xsens, Rokoko) have gained popularity. Those are used to approximate poses from body-mounted sensors using local measurements of acceleration and angular velocity, without external devices. However, IMUs are vulnerable to drift, require frequent recalibration, and struggle with long-term stability, error accumulation, and global positioning. Recently, a growing trend has been to reduce the number of mounted sensors, where sparse inertial methods have attempted to infer full-body motion from a limited number of IMUs~\citep{Yi:2022,Jiang:2022,Huang:2018,Yi:2021,Ponton:2023:SparsePoser}, but they still rely on relative motion and pre-calibrated kinematic models, often tailored to human morphology.

In this work, we propose a fundamentally different approach to sparse motion capture, that is intrinsic, scalable, and shape-invariant. Instead of relying on visual markers or inertial cues, we reconstruct articulated motion directly from pairwise distance measurements (PWD) between a sparse set of body-mounted wireless sensors. Using ultra-wideband (UWB) time-of-flight (ToF) rangings~\citep{Fontana:2004}, our method is resilient to lighting conditions and environmental complexity, and can operate reliably in unconstrained outdoor and non-laboratory settings. It facilitates simple installation, no body measurements are needed, and allows continuous and accurate tracking of any large articulated subject. This allows to expand the capturing possibilities even to animals (see~\cref{fig:teaser}).

Motion capture using UWB technology represents a relatively novel approach. Its potential lies in high frame rates and low latency, but it still faces several challenges such as noise, nonline-of-sight (NLoS) occlusions, and signal multipath interference. Recently, \citet{Armani:2024:UIP} explored this direction by integrating UWB with PWD estimation to enhance motion tracking, while employing an Extended Kalman Filter (EKF) to address measurement noise and sensor fusion. Their findings indicate that UWB and PWD offer a promising pathway for motion capture, particularly in complementing inertial systems. However, their method still relies primarily on IMU signals for reconstructing motion, with UWB-based positioning used to refine global accuracy.

To this end, we introduce \textbf{Wi}ld\textbf{P}oser (or \model for short), a real-time Transformer-based architecture that \review{relies solely on PWD measurements to predict full-body 3D joint positions and global motion, enabling motion capture in outdoor environments without requiring cameras or prior knowledge of the subject’s model.} The model follows a generative auto-regressive formulation, where the output represents a refined transformation of the input rather than a temporal continuation. \review{It can produce end-effector positions for any articulated character, supporting full-body articulation reconstruction across a wide range of subjects, and enabling the creation of animation mechanisms based on inverse kinematics (IK) to infer corresponding joint rotations (e.g., in animals). When prior knowledge of the character’s morphology is available (e.g., humans), learned methods can be applied to estimate joint rotations (e.g.,~\citep{Zuo:2021:Sparsefusion}), while in cases where inferred rotational information is available, a parametric 3D human body model, such as SMPL, can be directly reconstructed (e.g., via~\citep{Ponton:2025:DragPoser}).}
By enabling markerless motion capture of arbitrary articulated subjects in unconstrained environments, our system represents a key step toward capturing humans or even animals in the wild. It delivers smooth high-fidelity motion at $\sim$50 FPS with sub-60mm joint error, rivaling optical systems and outperforming inertial baselines. Our contributions include:

\begin{itemize}
    \item A portable mocap methodology that reconstructs 3D poses and global translation using only pairwise distance measurements. \review{Our approach eliminates the need for cameras, external infrastructure, or pre-calibrated kinematic models.} %
    \item A robust, shape-invariant system that generalizes across different body types and species, with a denoising mechanism specifically designed to mitigate noise and line-of-sight issues in UWB signals, enabling accurate motion capture in challenging environments. \review{When skeletal morphologies are known a priori, learned models can be integrated to produce accurate joint rotations suitable for character animation.}
    \review{\item A novel Refinement-Generative Model for motion capture, which takes a corrupted measurement and refines it given past predictions.}
\end{itemize}

\review{To support our claims, we conduct} extensive experiments, including in-the-wild recordings of humans and animals, highlighting the robustness and practical potential of our approach. \review{Comprehensive evaluations and ablation studies validate our hypotheses and highlight the superiority of our method, which achieves competitive or even superior performance compared to previous inertial-based methods. Our method shows strong potential for motion capture in \emph{outdoor} and unconstrained environments, opening new directions for future research in this area. Furthermore, it generalizes robustly across varying body morphologies and species, producing results comparable to contemporary optical-based systems.} %

\section{Related Work}

\subsection{Motion Capture Systems}

Mocap technologies, widely used in entertainment, healthcare, sports, and robotics, can be broadly classified into \emph{out-in} and \emph{in-out} systems. In \emph{out-in} systems, external tracking devices observe the subject from outside, typically using multiple cameras placed around a capture volume. These include \emph{passive systems} (e.g., Vicon, OptiTrack) that use retroreflective markers, and \emph{active systems} (e.g., PhaseSpace, Qualisys) that employ uniquely identifiable LEDs. Such systems offer millimeter-level precision via multi-angle triangulation, and high sample rates. However, they are limited to controlled indoor environments, being sensitive on changes in lighting conditions, require precise calibration, and are costly. Occlusions, marker swapping, and data cleaning remain challenges~\citep{Peng:2015,Aristidou:2018:SSA,Yu:2007}, though recent deep learning methods improve data denoising and reconstruction~\cite{Holden:2018,Chen:2021:MocapSolver,Pan:2024:RoMo}. 

In contrast, \emph{In-out} systems use body-worn sensors to estimate motion relative to the world. \emph{IMU-based systems} (e.g., Xsens, Rokoko) measure rotational data via gyroscopes, accelerometers, and magnetometers, integrating sensor fusion and biomechanical modeling to infer joint poses. These systems are portable, self-contained, cost efficient, and functional outdoors. However, they are shape-specific, requiring calibration to the subject's body and relying on forward kinematics to apply the computed rotations, they suffer from lower positional accuracy, orientation drift that accumulates over time, and the lack of a global reference (unless supplemented with GPS or external beacons)~\cite{Ami-Williams:2024}. Recent efforts aim to reconstruct full-body pose using fewer IMUs (e.g., Vive Tracker, Sony Mocapi), utilizing IK, learned motion priors, and machine learning models. Sparse sensor setups continue to improve via deep learning~\cite{Huang:2018,Yi:2021,Winkler:2022,Jiang:2022}, and physics-based methods~\cite{Yi:2022}. Hybrid solutions combining IMUs with monocular cameras and SLAM mitigate drift~\cite{Guzov:2021,Yi:2023}, while 6-DoF virtual reality devices enhance global pose estimation using autoencoders~\cite{Ponton:2023:SparsePoser}, transformers~\cite{Jiang:2022b,Zheng:2023}, and generative approaches~\cite{Du:2023,Castillo:2023}. Despite advancements, they still face challenges related to sparse input variability and constraint flexibility~\cite{Lee:2023}. Latent space models, including variational autoencoders and motion priors~\cite{Ling:2020,Rempe:2021}, enable efficient synthesis and reinforcement learning~\cite{Peng:2022}, while direct latent optimization and learned IK~\cite{Ponton:2025:DragPoser} have improved frame-level accuracy. Yet, issues persist in fine-grained control, contact point fidelity, and real-time detail reproduction~\cite{Shi:2023}. 

An emerging category of \emph{in-out} tracking includes cameras mounted on the body or headgear (e.g., Meta Quest) to capture the surrounding environment and infer user motion and position~\cite{Jiang:2024:Egoposer,Hollidt:2024,Yin:2024,Millerdurai:2024}. Some others also infer IMU data from wearable devices~\citep{Lee:2024:MocapEvery} to improve accuracy. While these \emph{egocentric} systems offer increased portability and are infrastructure-free, they face several limitations, including limited field of view, self-occlusions that hinder full-body pose estimation, motion blur, and sensitivity to camera placement. 

In summary, despite advancements, \emph{in-out} systems continue to struggle with global localization, exhibiting positional drift and lacking a reliable global reference without external aids.
\vspace{-1em}

\subsection{Ultra-Wideband Ranging and Position Reconstruction}

Ultra-wideband (UWB) is well-suited for high-precision localization due to its low-power, short-duration pulses and robustness to interference~\citep{Zafari2019, Delamare2020, Siwiak2004}. It estimates distances via Time-Difference-of-Arrival (TDoA)~\citep{Gustafsson:2005} and has been applied to tracking, robotics, and AR~\citep{Alarifi:2016, Lindhe:2006, Lymberopoulos:2015}. However, dynamic environments and NLoS conditions introduce significant error~\citep{Mok2012}, often addressed via impulse response analysis~\citep{Barral:2019}, Kalman-based sensor fusion~\citep{Piltaver2015, Tsaregorodtsev2019}, or statistical modeling~\citep{Hamie2015, Kulikov2020}.

Prior work has also investigated body-mounted UWB anchors to enable peer-to-peer localization without reliance on fixed infrastructure~\citep{DiRenzo2007, Hamie2013, Guizar2016}. To address challenges such as signal degradation due to body shadowing, probabilistic methods like particle or Kalman filtering have been employed~\citep{Tian2019, Otim2020}. Furthermore, combining UWB measurements with IMUs has been shown to enhance pose estimation~\citep{Corrales2008, Zihajehzadeh2017}, while more recently, UWB has been integrated to stabilize IMU-based mocap systems~\citep{Armani:2024:UIP}.

Our method addresses a key gap in prior work: the lack of a portable, full-body mocap system that utilizes recent advances in UWB technology and machine learning for pose reconstruction and denoising. \review{The most related systems to ours are Ultra Inertial Poser (UIP)~\citep{Armani:2024:UIP} and UMotion~\citep{Liu:2025:UMotion}, which utilize IMU measurements and UWD for PWD matrices. UIP estimates 3D pose and global translation using IMUs, with UWB pairwise distances used only for stabilization and refinement, while denoising is handled through EKF-based fusion. UMotion is a state estimation framework that employs an Unscented Kalman Filter along with an LSTM for pose and per-joint estimation, while omitting global translation.} %
In contrast, our method relies solely on inter-sensor PWDs, using transformers to infer sensor positions and reconstruct full-body poses and global translation. We introduce a novel denoising model to address noise and NLoS challenges, with optional IMU integration to further improve accuracy. Unlike traditional IMU-based systems, our approach is infrastructure-free, shape-invariant, and applicable across different body types and species, offering a convenient and portable mocap system.

\section{Methodology}

\subsection{Preliminaries}
Common practices of sparse mocap involve the use of up to six IMUs placed on keypoint joints to estimate pose and global translation~\citep{Ponton:2023:SparsePoser,Yi:2022}. %
Those typically produce joint hierarchies based on SMPL~\citep{Loper:2015:SMPL}. However, we tackle this task differently. Instead of approximating partial poses from inertial data, we introduce \textit{\modelNoSpace}, a Transformer-based architecture that reconstructs 3D poses from sparse, noisy PWD measured by six UWB sensors. This section defines our model group, followed by the architecture, training procedure, and loss functions.

\begin{figure*}[htbp]
    \centering
    \includegraphics[width=0.95\textwidth]{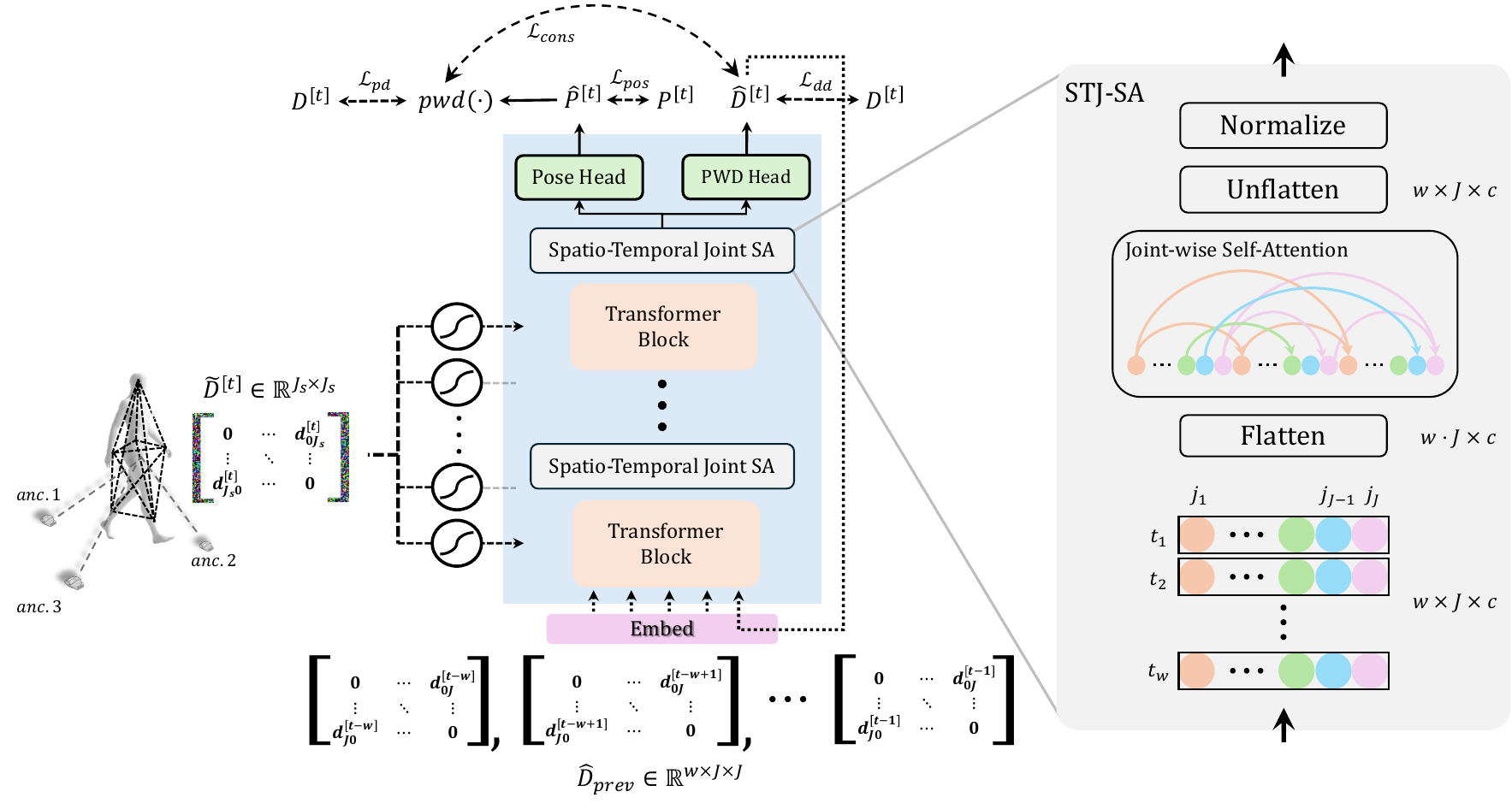}
    \vspace{-0.5em}
    \caption{\textit{Left}: \model is a Motion Capture Refinement-Generative Model (see \cref{def:2}). Given a clean sequence of pairwise distances $\hat{D}_{prev} \in \mathbb{R}^{w \times J \times J}$ up to timestamp $t-1$, and a \emph{sparse} noisy measurement $\tilde{D}^{[t]} \in \mathbb{R}^{J_{\text{s}} \times J_{\text{s}}}$, the model predicts the next skeleton $\hat{P}^{[t]} \in \mathbb{R}^{J \times 3}$ in global coordinates. Simultaneously, it predicts the denoised pairwise distance matrix $\hat{D}^{[t]} \in \mathbb{R}^{J \times J}$ for a timestamp $t$, which is later used as input to predict pose $t+1$. To ensure consistency between the two output modalities, we apply a consistency loss $\mathcal{L}_{\text{cons}}$, in addition to separate reconstruction losses $\mathcal{L}_{\text{pd}}$ and $\mathcal{L}_{\text{dd}}$ for each modality. \textit{Right}: The STJ-SA layers, inserted at the end of each Transformer block, enable the model to learn correspondences between joints across time, aligning per-joint trajectories and mitigating jitter caused by noisy measurements. %
    We use $J_s = 6$ for the sparse configuration and $J = 24$ for the full SMPL skeleton representation.}
    \label{fig:model}
    \vspace{-1em}
\end{figure*}

\subsection{Problem Formulation}

\begin{definition}[Refinement-Generative Model]
A \emph{Refinement-Generative Model} $\mathcal{M}_\theta$ is defined as a mapping:
\begin{equation}
    \mathcal{M}_\theta: \mathcal{X} \times \tilde{\mathcal{Y}} \rightarrow \mathcal{Y},
\end{equation}
where: $\mathcal{X}$ denotes the input features (e.g., past predictions of a sequence model), $\tilde{\mathcal{Y}}$ represents a noisy or corrupted version of the target output (e.g., an auxiliary or intermediate signal), and $\mathcal{Y}$ is the clean target output to be reconstructed.

The model is termed \emph{refinement-generative} because it leverages $\tilde{\mathcal{Y}}$, a degraded form of the true target $\mathcal{Y}$, as an auxiliary signal. Conceptually, $\mathcal{M}_\theta$ can be interpreted as a \emph{Generative Noise Filter}, wherein previous context and noisy guidance jointly inform the generation of a denoised or corrected version of $\mathcal{Y}$, conditioned by prior information.
\end{definition}

\begin{definition}[Motion Capture Refinement-Generative Model]\label{def:2}
Let $P^{[t]} \in \mathbb{R}^{J \times 3}$ denote an ordered 3D pointset of $J$ joints at timestamp $t$, representing the spatial coordinates of a skeleton pose. Additionally, let $D^{[t]} \in \mathbb{R}^{J \times J}$ be the precomputed pairwise Euclidean distance matrix at the same timestamp, defined as:
\begin{equation}
    D^{[t]}_{ij} = \|P^{[t]}_i - P^{[t]}_j\|_2,
\end{equation}
for all $i, j \in \{1, 2, \dots, J\}$.

A \emph{Motion Capture Refinement-Generative Model} is a particular instance of the refinement-generative model $\mathcal{M}_\theta$, taking as input:
\begin{itemize}
    \item A temporal window $w$ of prior poses $\mathcal{X} = \left\{P^{[t-w]}, \dots, P^{[t-1]}\right\}$,
    \item The current noisy observation $\tilde{\mathcal{Y}} = D^{[t]}$,
\end{itemize}
and outputs an estimate of the current 3D pose:
\begin{equation}
    \mathcal{M}_\theta(\mathcal{X}, \tilde{\mathcal{Y}}) = \hat{\mathcal{Y}} = \hat{P}^{[t]} \in \mathbb{R}^{J \times 3}.
\end{equation}

The primary objective is to reconstruct the pose $\hat{P}^{[t]}$ at timestamp $t$ based on the pairwise distance matrix $\tilde{\mathcal{Y}} = D^{[t]}$ and the sequence of previous poses $\mathcal{X}$. In this formulation:
\begin{itemize}
    \item $P^{[t]}_j$ denotes the 3D coordinates of joint $j$ at time $t$,
    \item $D^{[t]}_{ij}$ encodes the Euclidean distance between joints $i$ and $j$.
\end{itemize}
\end{definition}

\subsection{\model~Architecture}

\model is a Motion Capture Refinement-Generative Model (denoted as $\mathcal{RG}_\theta$), which combines concepts from Encoder-Decoder~\citep{Vaswani:2017:AttentionIsAllYouNeed,Raffel:2020} and Decoder-only Transformers~\citep{Radford:2019:Language,Touvron:2023:Llama2O}. Decoder-only models consist of $n$ stacked blocks, each composed of a masked multi-head self-attention operation followed by a feed-forward network, wrapped in residual connections and layer normalization. Unlike conventional approaches, we augment each block with a cross-attention mechanism, where the \emph{Keys} and \emph{Queries} are derived from the pairwise distance matrix $D^{[t]}$, and the \emph{Values} are obtained from the sequence of past poses $P_{\text{prev}}$. This interaction enables the model to dynamically balance temporal consistency and noisy input signals, and is achieved by two training stages. \cref{fig:model} provides a visual overview of the \model architecture. %

\subsubsection{Generation in the Pairwise-Distance Space}
We found that predicting the next pose in terms of its PWD representation was more effective. Therefore, \model~is trained to generate motion in the \textit{pairwise-distance space}. Specifically:
\begin{equation}
  \begin{aligned}
    h^{[t]} = \mathcal{RG}_\theta\left(\hat{D}_{\text{prev}}, \tilde{D}^{[t]}\right)
  \end{aligned}
\end{equation}
where $\hat{D}_{\text{prev}}\in\mathbb{R}^{w\times J \times J}$ are the $w$ previously predicted PWDs up to timestamp $t-1$, and $\tilde{D}^{[t]}\in\mathbb{R}^{J\times J}$ is the current noisy input. The hidden state $h^{[t]} \in \mathbb{R}^d$ is then passed through Pose and PWD Heads to yield the next pose $\hat{P}^{[t]} \in \mathbb{R}^{J\times 3}$ and the PWD $\hat{D}^{[t]} \in \mathbb{R}^{J\times J}$, correspondingly.
Next, to predict $\hat{P}^{[t+1]}$, $\hat{D}^{[t]}$ is embedded and concatenated with previous predictions.

\subsubsection{Rigid Body Generation}
\label{subsubsection:Rigid Body Generation}

While \model~is not inherently limited to human motion, most practical applications involve capturing humans. To this end, we introduce an extended variant, termed \emph{\modelNoSpace-H}, trained to reconstruct dense skeletal structures of the SMPL body model (where $J=24$). We denote $J_s$ and $J$ as the size of the sparse and dense joint sets, respectively.

Given a sparse input $\tilde{D}^{[t]} \in \mathbb{R}^{J_s \times J_s}$, the model predicts a dense pointset $\hat{P}^{[t]} \in \mathbb{R}^{J \times 3}$. To encourage learning of rigid-body constraints and relations, we incorporate a \textit{Rigidity Loss} term (see~\cref{sscec:reg_loss}). Since previous works focus on human motion, we use this variant in all our comparisons (consistent with prior work, we set $J_s = 6$ for the sparse setting).

\subsubsection{Global Displacement and Rotation} \label{sec:g_rot}

\begin{figure}[htbp]
    \centering
    \includegraphics[width=0.70\columnwidth]{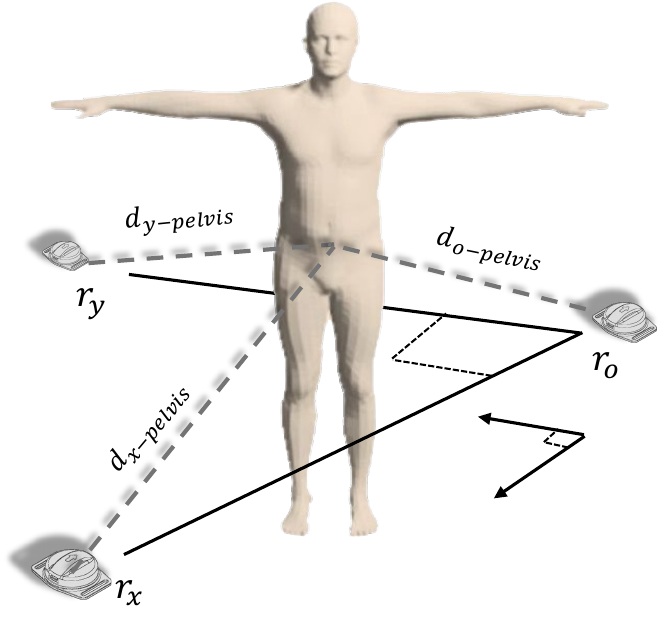}
    \caption{\textbf{Reference Anchors Placement}. To address global orientation of recorded subjects, we simulate a 2D global system by adding three sensors to the $\left|J_s\right|$-sized set. Mapping them to fixed spatial locations ($r_o \rightarrow \left(0,0,0\right), r_x \rightarrow \left(1,0,0\right)$, $r_y \rightarrow \left(0,1,0\right)$) and optimizing \cref{eq:pd_loss,eq:dd_loss,loss:gravity}, \model predicts precise global positions.}
    \label{fig:refs}
\end{figure}

Our mocap system does not require body calibration or predefined initial poses. While our pose reconstruction depends solely on PWD measurements between body-mounted sensors, we introduce three additional fixed reference sensors (anchors) placed on the ground to form an orthogonal 2D basis (see~\cref{fig:refs}). These anchors serve as a spatial reference, allowing the model to learn global orientation and position through supervision. We avoid the third axis to ease the setup, and allow more flexibility. These anchors are integrated into the decentralized sensor network across the subject's body. To mitigate reflection ambiguities in the XY-plane, we introduce a $\textit{Gravity Term}$ in our loss function, which constraints the predicted skeletons to remain above the ground surface. By incorporating reference sensor positions, the model consistently recovers global displacement and orientation across frames. More details in~\cref{ssec:orient_loss,ssec:gravity_reg}. We highlight that this component operates in parallel, and it is not involved in the pose reconstruction evaluation process. Comparable methods handle this challenge using Vive trackers and base stations~\citep{Ponton:2023:SparsePoser} or via approximation strategies.

\subsubsection{Gated Cross Attention (Gated CA)}

In standard encoder-decoder Transformer architectures~\citep{Raffel:2020,Vaswani:2017:AttentionIsAllYouNeed}, cross-attention allows context to guide predictions. In our refinement-generative setting, we must balance the amount of attention between the past pose predictions and the present noisy input measurement. To achieve this, inspired by~\citep{Alayrac:2022:Flamingo}, we replace traditional cross-attention with a learnable \emph{Gated Cross Attention}. For a layer $l$ we apply:
\begin{equation}
  \begin{aligned}
    A_{\text{ca}} &= \text{CA}_l\left(\text{Emb}\left(\tilde{D}^{[t]}\right), h^{[t]}_{l-1}\right), \\
    A_{\text{gate}} &= A_{\text{ca}} \odot \sigma\left(\text{W}_g h^{[t]}_{l-1}\right), \\
    h^{[t]}_{l} &= A_{\text{gate}} + \text{dropout}\left(h^{[t]}_{l-1}\right),
  \end{aligned}
\end{equation}
where $\text{CA}_l$ is a cross-attention in layer $l$, $h^{[t]}_{l-1}$ is the output of the previous layer $l-1$, and $\text{Emb}\left(\cdot\right)$ is an embedding operation. $\text{W}_g$ is a learnable gating matrix, $\sigma$ is the sigmoid function, and $\odot$ denotes the Hadamard product operation. Given that our input measurements are noisy, the gating mechanism allows the model to control the flow and dynamically attend them at every timestamp.

\subsubsection{WiP-SI (Shape-Invariant)} \label{sec:orient_mds}
Since \model is trained primarily on human motion data, it implicitly learns strong priors over human joint configurations. To improve generalization, we introduce \emph{\modelNoSpace-SI}, a variant trained to reconstruct sparse joints only (where $J=J_s$). To further support generalization to non-human morphologies (e.g., quadrupeds), we apply a random joint permutation at each training step. Let $\pi:{1,\dots,J} \rightarrow {1,\dots,J}$ be the sampled permutation, and $D^{[t]'}$ the resulting permuted matrix:
\begin{equation}
    \begin{aligned}
        D^{[t]'}_{ij} = D^{[t]'}_{ji} = D^{[t]}_{\pi(i)\pi(j)}.
    \end{aligned}
\end{equation}

This permutation acts as a regularizer, weakening the model's over-reliance on specific joint identities while preserving global spatial relations. We call this variant \modelNoSpace-SI, as it supports arbitrary joint sets without prior knowledge of the subject's shape. In our experiments, we apply this variant only to non-human subjects.

\subsubsection{PWD Head}
\model is designed to directly predict joint positions in the global coordinate space. However, this focus on global prediction can lead to the loss of important local structural information during training. To address this, we add an auxiliary \emph{PWD head}, which encourages the preservation of skeletal geometry. This head predicts PWDs between all joints, including anchors, of the output skeleton, helping the model maintain local consistency and better learn the spatial relationships between joints, complementing global position prediction. The predicted PWD $\hat{D}^{[t]}$ is then concatenated with $\hat{D}_{\text{prev}}$, to predict the next pose (see~\cref{fig:model}-left).

\subsubsection{Spatio-Temporal Joint Self-Attention (STJ-SA)}
The shift from clean to noisy PWD is not trivial and each training stage introduces different challenges. To maintain smooth and accurate motion, it is crucial for the model to emphasize joint-wise temporal coherence. To achieve this, we introduce an STJ-SA layer at the end of every Transformer block. The core idea of this layer is to enable the model to learn joint-wise relations across time. This is done by flattening the hidden state $h^{[t]}_l\in \mathbb{R}^{T\times J\cdot c}$ to a shape of $T\cdot J\times c$ before the SA operation, followed by post-normalization and un-flattening back (\cref{fig:model}-right). This modification enables dynamic smoothing across joints while preserving consistency on the original skeletal structure, thereby producing reconstructions that are both temporally stable and anatomically plausible.

\subsection{Losses}
To train our model, we define a unified objective composed of multiple loss terms. Each term contributes to an equilibrium that balances temporal coherence, anatomical plausibility, and robustness to noise, while minimizing reconstruction errors in the PWD space. Throughout this section, we denote the Mean Squared Error as $\text{MSE}$.

\subsubsection{PWD Losses}

A key goal of our model is to preserve skeletal structure, for which we define the following loss components:
\begin{equation}
  \mathcal{L}_{\text{pd}} = \text{MSE}\left(\text{pwd}\left(\hat{P}^{[t]}\right), D^{[t]}\right),
  \label{eq:pd_loss}
\end{equation}
\begin{equation}
  \mathcal{L}_{\text{dd}} = \text{MSE}\left(\hat{D}^{[t]}, D^{[t]}\right),
  \label{eq:dd_loss}
\end{equation}
\begin{equation}
  \mathcal{L}_{\text{cons}} = \text{MSE}\left(\text{pwd}\left(\hat{P}^{[t]}\right), \hat{D}^{[t]}\right).
  \label{eq:consistency_loss}
\end{equation}

Here, $\text{pwd}(\cdot)$ computes the full pairwise Euclidean distance matrix of a pointset in any $n$-dimensional space. $\mathcal{L}_{\text{pd}}$ (\cref{eq:pd_loss}) penalizes errors in skeletal distances from predicted 3D joints, while $\mathcal{L}_{\text{dd}}$ (\cref{eq:dd_loss}) directly supervises the predicted pairwise distances. The consistency loss $\mathcal{L}_{\text{cons}}$ (\cref{eq:consistency_loss}) aligns the two output heads, enforcing internal coherence between predicted poses and their distance representations. Together, these losses ensure both anatomical consistency and geometric fidelity.

\subsubsection{Orientation Loss} \label{ssec:orient_loss}

As introduced in ~\cref{sec:g_rot}, we use three fixed reference sensors on the ground (denoted $r_{\{1,2,3\}}$) to define a global orientation basis. By mapping these to fixed 3D locations, we implicitly anchor the model's global rotation during training. Additionally, we supervise their predicted positions by applying:
\begin{equation}
  \mathcal{L}_{\text{refs}} = \sum_t\left\Vert r_i^{[t]} - \hat{r}_i^{[t]}\right\Vert_2,
\end{equation}
where $r_i^{[t]}$ and $\hat{r}_i^{[t]}$ are the ground-truth and predicted positions of reference sensor $i$ at time $t$, respectively.

\subsubsection{Gravity Regularization} \label{ssec:gravity_reg}

While the orientation anchors provide global rotation, they do not resolve reflection ambiguity about the XY-plane. To address this, we introduce a \textit{Gravity Term} that constrains joint positions to lie above the ground plane:
\begin{equation}
  \mathcal{R}_{\text{gravity}} = \mathbb{E}_{\hat{P}^{[t]}}\left[\max\left(0, -\hat{p}_{iz}^{[t]}\right)\right]
  \label{loss:gravity},
\end{equation}
where $\hat{p}_{iz}^{[t]}$ is the $z$-coordinate of joint $i$ in the predicted pose $\hat{P}^{[t]}$. This regularizer penalizes any joint predicted below the ground surface, enforcing a correct global orientation with respect to gravity.

\subsubsection{Velocity Loss.} \label{ssec:velo_loss} Inspired by prior work ~\citep{Tevet:2023:MDM}, %
we enforce additional supervision on the joints' velocity to ensure smooth and reliable transitions between consecutive timestamps:
\begin{equation}
    \mathcal{L}_{\text{velo}} = \sum_{t} \sum_{j} \left\Vert \left(\hat{P}^{[t]}_j - \hat{P}^{[t-1]}_{j}\right) - \left(P^{[t]}_j - P^{[t-1]}_{j}\right)
    \right\Vert_2.
\end{equation}
This loss tracks both the velocity magnitude, and directional velocity to avoid jittering. This loss is added in the second training stage.
\subsubsection{Rigidity Loss} \label{sscec:reg_loss}

When training the model to predict the full SMPL skeleton (24 joints), we introduce a \textit{rigidity constraint} to enforce constant bone lengths between connected joints. This encourages preservation of the articulated body structure by applying:
\begin{equation}
  \mathcal{L}_{\text{rigidity}} = \sum_{(i,j) \in \mathcal{B}}\left\Vert b_{ij} - \hat{b}_{ij}\right\Vert_2,
\end{equation}
where $\mathcal{B}$ is the set of rigid bone pairs in the SMPL model. The ground-truth and predicted bone lengths are easily obtained as $b_{ij} = D^{[t]}_{ij}$ and $\hat{b}_{ij} = \hat{D}^{[t]}_{ij}$ for each bone $(i, j) \in \mathcal{B}$, respectively.

To this end, the overall objective of our training process is:
\begin{equation}
    \operatorname*{argmin}_{\theta\in WiP}\sum_i \lambda_i\mathcal{L}_i,
\end{equation}
where $\lambda_{i}$ is a predefined weight for each loss term $i$.

\subsection{Noise Handling}

UWB signals offer high temporal resolution and robustness to interference, making them well-suited for localization tasks. However, they remain inherently vulnerable to noise, occlusions, and NLoS conditions, factors that can introduce significant motion reconstruction errors (with an up to 15~cm drifts as reported by manufacturers). To address these challenges, we introduce a secondary training stage, referred to as the \emph{denoising phase}. In this phase, the model is fine-tuned on the original dataset augmented with simulated noise that reflects real-world measurement distortions. We define a noise function that simulates raw noise, yet remains simple and general:
\begin{equation}
    \phi\left(D^{[t]}\right) = \frac{1}{w} \sum_{i = t-\lfloor{w/2}\rfloor}^{t + \lfloor{w/2}\rfloor} \left( D^{[i]} + \epsilon_i \right), \quad \epsilon_i \sim \mathcal{N}(0, \sigma^2)\in\mathbb{R}^{J_s\times J_s}.
  \label{eq:noise_fn}
\end{equation}

We found that setting $\sigma=15$~cm and $w=5$ achieves decent approximation of the measured noise. This augmentation strategy enhances generalization while preventing overfitting to specific capture instances.

\section{Experiments}

\subsection{Experimental Setup}
\subsubsection{Implementation Details} \label{sec:imp_det}

We train our model for 20 epochs (in both training stages) on 4 NVIDIA GeForce RTX 3090 using DDP training in PyTorch. For UWB-based PWD recordings, we employ DWTAG100 sensors~\citep{CiholasDWTAG}, and use a maximum frequency of $20$ Hz. More details can be found \review{in~\Cref{sec:Hardware Details}}.%

\subsubsection{Training Procedure} To train \model to handle raw measurements of PWD, we divide the training into two stages: i) the \textit{distance-to-motion phase}, where the model is exposed to clean distance data and trained to produce continuous and accurate motion, and ii) the \textit{denoising phase}, where we synthetically add noise in the distance space (\cref{eq:noise_fn}), to corrupt the input data. In this stage, the model is trained to handle noisy inputs, while still producing plausible motion sequences. In the second stage,  we freeze $95\%$ of the model's parameters, and train only the Gated Attention and the newly added STJ-SA layers. The key motivation behind this approach is to preserve the model's motion reconstruction accuracy while improving its resilience to degraded and noisy input data.

\subsubsection{Data Prepossessing} For training, we employ the publicly available motion dataset AMASS~\citep{Mahmood:2019:AMASS}. Unlike previous approaches, we train \model to reconstruct a prefixed 3D pointset, which forms a set of joints in global world. Therefore, %
we first convert the SMPL representation to a set of $24$ joints and create sequences of motion. In all our experiments, we set the window length at $16$ ($\sim$250 ms). Then, we normalize the head-pelvis distance to be $1$ across all sessions, and align the floor with the $XY$-plane.

\subsubsection{Data Collection for Evaluation}
\label{subsubsection: Data Collection}
To collect evaluation data for our system, we attach six UWB sensors to a human subject: at the pelvis, head, hands, and feet. \review{In contrast to prior approaches that position sensors on the knees~\citep{Armani:2024:UIP,Yi:2022}, we attach sensors to the feet, similarly to \citet{Ponton:2023:SparsePoser}. The feet, as natural end-effectors and primary contact points with the environment, enable more precise end-effector localization, provide longer inter-sensor distances, and experience fewer NLOS issues. This configuration supports the ongoing trend toward sparse, body-mounted sensing for motion capture. Although additional sensors could improve accuracy, shorter inter-sensor distances may increase noise sensitivity due to UWB limitations at close range.} To obtain precise ground-truth motion data, we simultaneously recorded the 3D positions of the UWB sensors using a synchronized optical mocap system (PhaseSpace Impulse X2E), ensuring a well-aligned paired dataset.
We collected a total of $\sim$20 minutes length of $10$ different actions. %

\subsection{Results}
Our system was tested in diverse settings, including full-body human pose reconstruction from sparse PWD measurements, hybrid PWD+IMU inputs, and non-human motion capture in uncontrolled environments. The results highlight the versatility and robustness of our approach in various input modalities and subjects. For animated examples, please refer to the supplementary video.

\subsubsection{Full-Body Joint Position Reconstruction}
\begin{figure}[h]
    \centering
    \includegraphics[width=1\columnwidth]{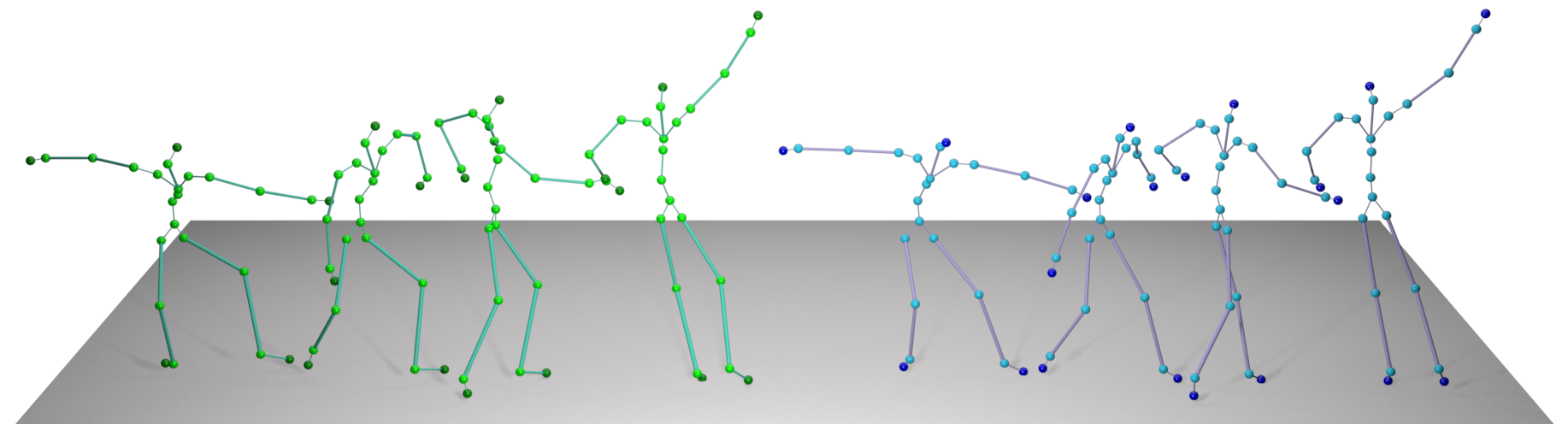}
    \caption{\textbf{Full-Body Reconstruction} (from DanceDB). Our method (\textcolor{darkblue}{blue}), relying solely on PWD measurements, reconstructs full-body joint positions over time that are consistent with the ground truth (\textcolor{darkgreen}{green}).}
    \label{fig:results-joints}
\end{figure}

Our approach reconstructs the full set of $J = 24$ joint positions using only PWD measurements from $J_s = 6$ body-mounted UWB sensors. Despite the limited spatial coverage, it effectively captures the underlying motion dynamics by using learned priors to produce anatomically consistent full-body joint positions. \cref{fig:results-joints} shows representative reconstructions compared to ground truth.

\subsubsection{\model for Character Animation}
\begin{figure}[h]
    \centering
    \includegraphics[width=1\columnwidth]{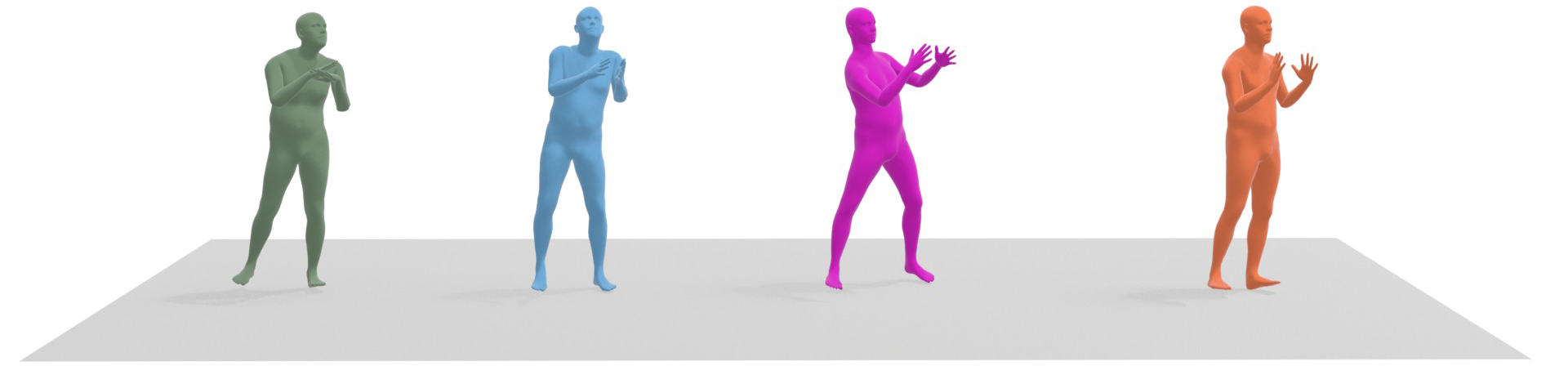}
    \includegraphics[width=1\columnwidth]{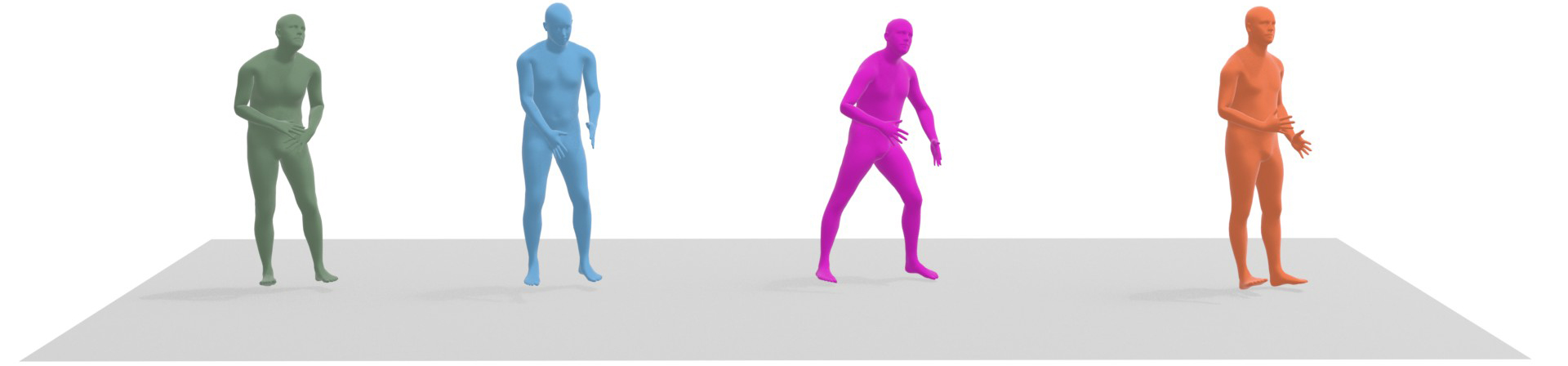}
    \includegraphics[width=1\columnwidth]{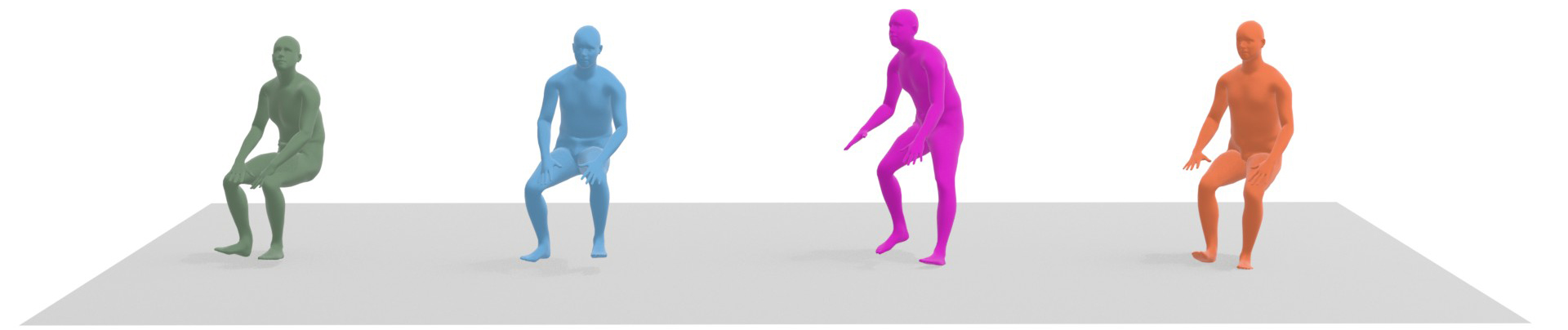}
    \includegraphics[width=1\columnwidth]{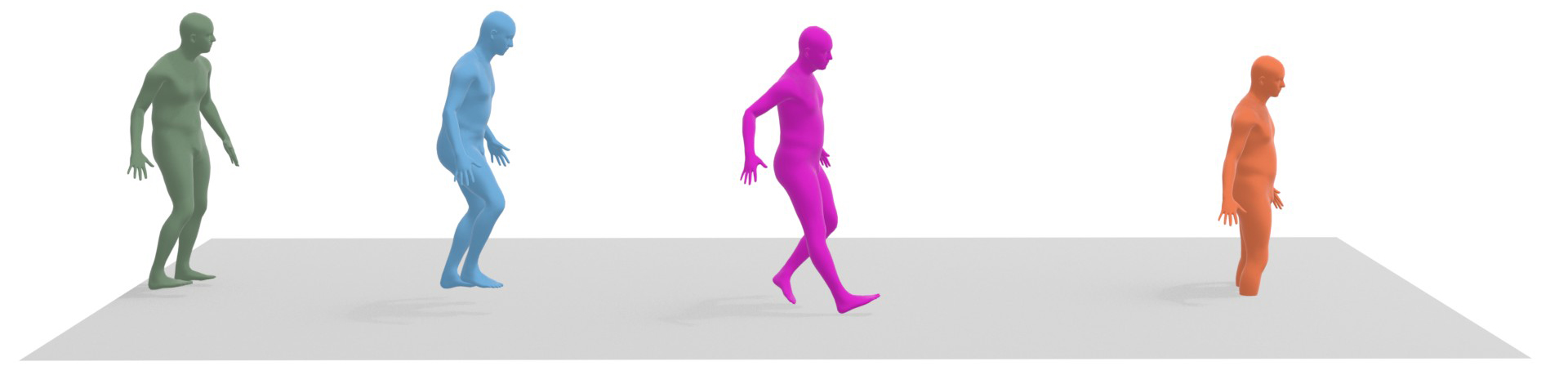}
    \caption{\review{\textbf{Qualitative Comparison on UIP-DB.} By integrating SMPLify3D, we achieve strong performance, even compared to methods that predict joint rotations directly. \textcolor{darkblue}{Blue} is our method, \textcolor{magenta}{magenta} is UIP, \textcolor{orange}{orange} is UMotion, and ground truth is in \textcolor{darkgreen}{green}.}}
    \label{fig:results-pose}
\end{figure}

\review{By design, \model predicts keypoint positions in 3D space. While effective, this representation is limited in its ability to capture local twisting articulations. To address this limitation, we extend our pipeline to support full-body virtual character animation by estimating SMPL pose parameters from 3D joints. Specifically, we employ the initial fitting used \mbox{by \citet{Zuo:2021:Sparsefusion}} (sec. 3.1), \emph{SMPLify3D}. This method estimates SMPL pose parameters from 3D joint positions by optimizing the rotation-based hierarchical representation with respect to learned pose priors. By integrating the above, we are able to recover both joint rotations and global motion (see \cref{fig:results-pose}).}

\subsubsection{SMPL Pose Estimation from PWD + Rotation Measurements}
\begin{figure}[h]
    \centering
    \includegraphics[width=1\columnwidth]{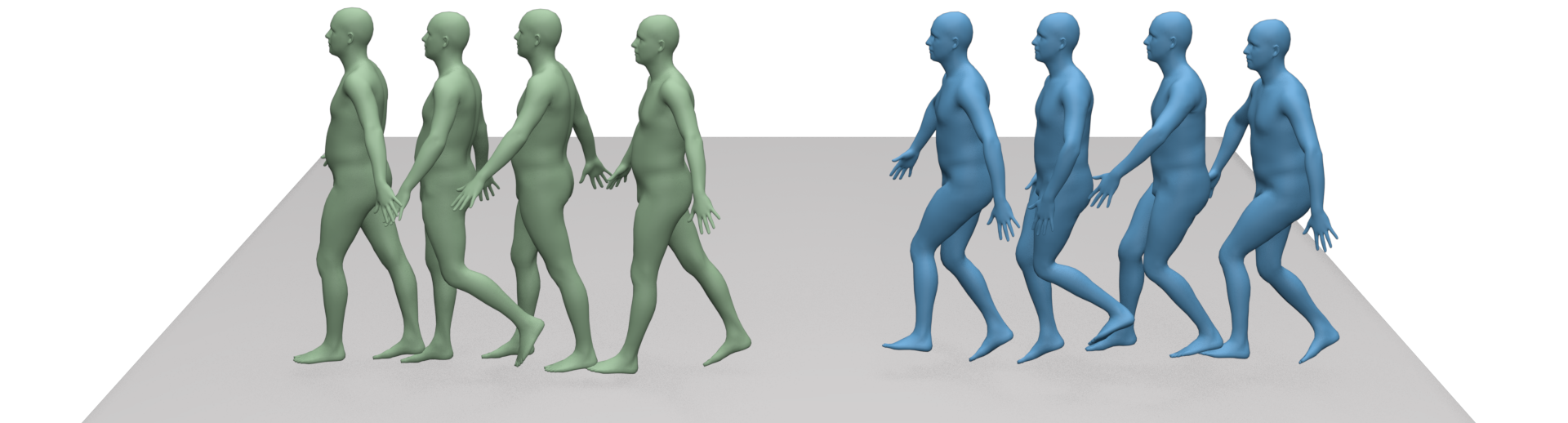}
    \caption{\textbf{SMPL Pose Estimation (from TotalCapture).} When both PWD and inertial data are available, our method (\textcolor{darkblue}{blue}) produces SMPL poses that closely match the ground truth (\textcolor{darkgreen}{green}).}
    \label{fig:results-SMPL}
\end{figure}

In scenarios where rotation data are available, we extend our pipeline to predict the full-body pose parameters of the SMPL model directly. By integrating DragPoser with our Transformer-based model, we accurately recover joint rotations and global motion. \cref{fig:results-SMPL} shows our reconstructions that closely match the ground-truth poses.

\subsubsection{Wild Capturing in the Wild}

\begin{figure}[htbp]
    \centering 
    \includegraphics[width=\columnwidth]{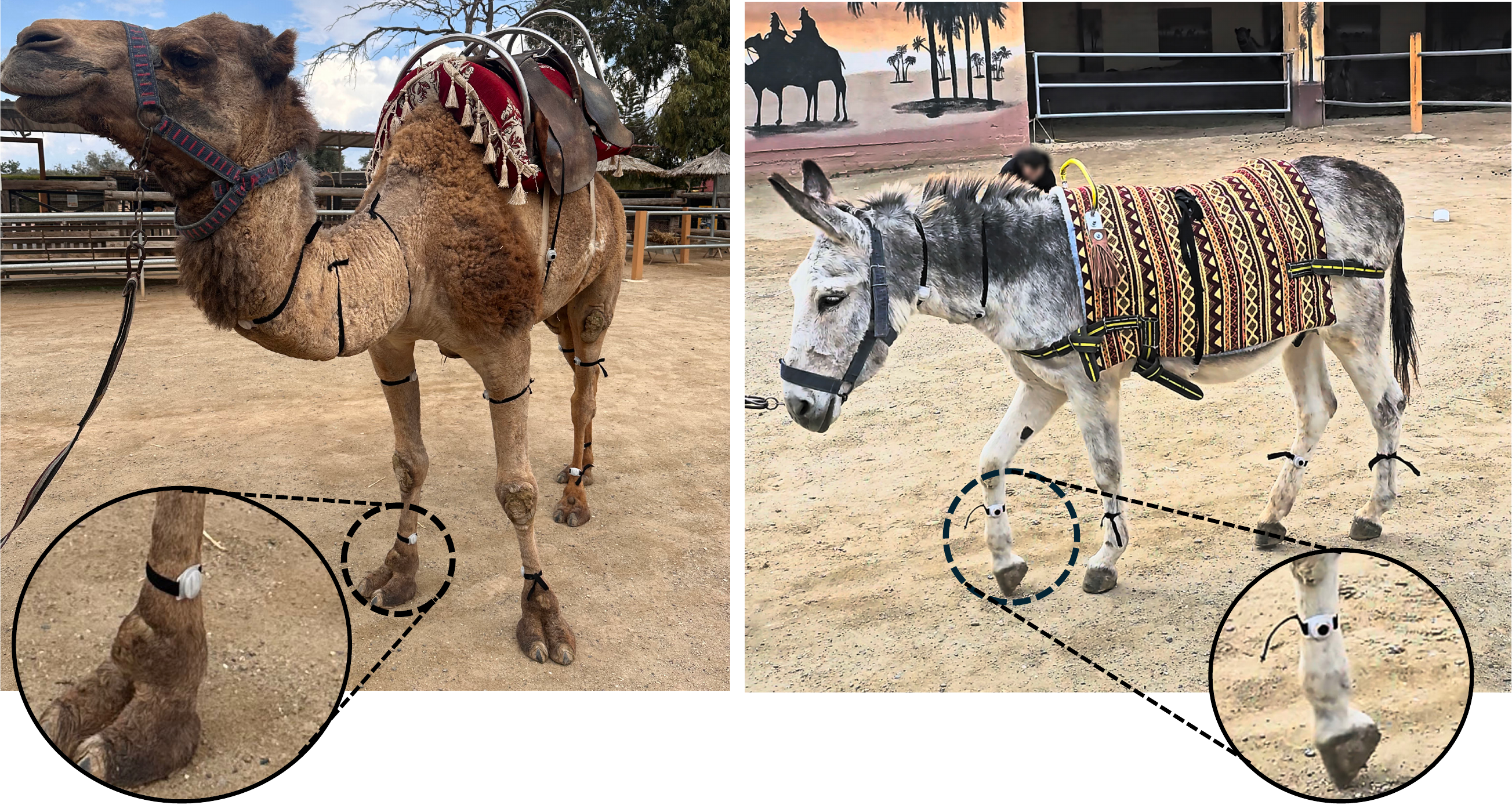}
    \caption{\textbf{Motion Capture on Wildlife.} Our methodology is suitable for any type of moving subject and environment. The UWB sensors can be easily attached to different animals (e.g., camels, donkeys), and thanks to our body-calibration-free approach, we can capture their motions.} 
    \label{fig:animals} 
\end{figure}

To demonstrate the generality of our approach across species, we attached multiple UWB sensors to a camel and a donkey (see~\cref{fig:animals}). Due to limited data for training a pose model based on learning, we used the \mbox{\emph{\modelNoSpace-SI}} variant to reconstruct the global 3D positions of the sensor. For animation, we built constrained, rigged 3D models and applied IK, treating the reconstructed positions as end-effectors that drive the motion of the limb and torso. To improve motion fidelity, we segmented the cyclic gait patterns of the animals and computed a temporal average across gait cycles, reducing noise, anomalies, and outliers in the raw data. In addition, we enforced foot contact constraints with the ground to enhance physical plausibility and stability during locomotion. Secondary body parts such as the head, tail, and ears remain static, as no sensors were mounted in these regions. \cref{fig:teaser} presents selected snapshots from our animated sequences\review{, while full videos are included in the supplementary materials}. 

\section{Evaluation and Discussion}

In this section, we analyze the performance of our method in comparison to existing baselines and state-of-the-art approaches. We also conducted an ablation study to evaluate the contribution of \modelNoSpace's architectural components, demonstrating how individual modules contribute to the robustness of the reconstructed motion. Please refer to \review{\Cref{sec:Appendices,sec:Further Implementation Details}} for additional implementation details, as well as extended qualitative and quantitative results.
\begin{table*}[ht]
\centering
\caption{\textbf{Comparison with previous Sparse-sensor mocap methods}. When evaluated on synthetically generated data, \model outperforms prior methods, even thought they utilize multiple measurement modalities, including \textbf{O}rientation, \textbf{A}cceleration, angular \textbf{V}elocity, and \textbf{D}istances. Notably, our approach does not require body-specific calibration or an initial pose for evaluation. \review{Bold indicates the best and underline the second best performance per metric.}} %
\vspace{-0.5em}
\label{tab:comparisons}
\footnotesize
\def\arraystretch{1.2}
\resizebox{\textwidth}{!}{
\begin{tabular}{lccccccclcccccccccc}
\hline
\multirow{2}{*}{Method} & \multirow{2}{*}{\begin{tabular}[c]{@{}c@{}}Inp. Measur. \\ Type\end{tabular}} & \multirow{2}{*}{\begin{tabular}[c]{@{}c@{}}Body Calib.\\ Free\end{tabular}} & \multicolumn{5}{c}{TotalCapture}                                                       &                      & \multicolumn{4}{c}{DIP-IMU}                                          & \multicolumn{1}{l}{} & \multicolumn{5}{c}{DanceDB}                                                            \\ \cline{4-8} \cline{10-13} \cline{15-19} 
                        &                                                                               &                                                                             & \scriptsize PE & \scriptsize EEE & \scriptsize GTE & \scriptsize AJE & \scriptsize GSE &                      & \scriptsize PE & \scriptsize EEE & \scriptsize AJE & \scriptsize GSE & \multicolumn{1}{l}{} & \scriptsize PE & \scriptsize EEE & \scriptsize GTE & \scriptsize AJE & \scriptsize GSE \\ \hline
TransPose               & O+A+V                                                                         & \xmark                                                                      & 5.57           & 9.07            & 8.99            & 0.58            & 2.35            &                      & 7.62           & 7.43            & 1.60            & 2.06            &                      & 8.50           & 11.41           & 6.68            & 1.35            & 3.61            \\
TIP                     & O+A+V                                                                         & \xmark                                                                      & 5.14           & -               & -               & 0.35            & -               &                      & 5.71           & 7.63            & 1.31            & 2.22            &                      & 8.46           & 11.66           & 12.48           & 2.61            & 3.52            \\
PIP                     & O+A+V                                                                         & \xmark                                                                      & 5.61           & 9.17            & 8.55            & 0.25            & 2.39            &                      & 5.04           & 7.67            & 0.40            & 2.06            &                      & 5.69           & 8.55            & 9.44            & 1.81            & {\ul 2.50}      \\
UIP                     & O+A+V+D                                                                       & \xmark                                                                      & 5.49           & 10.25           & 24.70           & 0.32            & \textbf{1.52}   &                      & 5.05           & 5.59            & 0.46            & {\ul 1.58}      &                      & 7.61           & 13.19           & 20.94           & 1.05            & 2.53            \\
UMotion                 & O+A+V+D                                                                       & \xmark                                                                      & 4.46           & 6.74            & -               & 0.20            & 1.72            &                      & \textbf{2.81}  & 5.15            & 0.39            & \textbf{1.31}   &                      &                &                 & -               &                 &                 \\ \hline
\modelNoSpace-Geo       & D                                                                             & \cmark                                                                      & 5.61           & {\ul 6.30}      & {\ul 4.82}      & 0.12            & 3.04            & \multicolumn{1}{c}{} & 3.50           & {\ul 2.71}      & 0.24            & 1.84            &                      & 4.95           & {\ul 4.30}      & 2.70            & 0.64            & 2.61            \\
\modelNoSpace-knees     & D                                                                             & \cmark                                                                      & \textbf{3.07}  & 6.91            & 4.98            & {\ul 0.10}      & 2.11            & \multicolumn{1}{c}{} & {\ul 3.40}     & 5.86            & \textbf{0.23}   & 2.24            &                      & {\ul 4.79}     & 8.43            & \textbf{2.28}   & {\ul 0.61}      & 3.21            \\
\model (ours)           & D                                                                             & \cmark                                                                      & {\ul 4.17}     & \textbf{4.18}   & \textbf{1.66}   & \textbf{0.09}   & 2.71            &                      & 3.80           & \textbf{2.14}   & {\ul 0.24}      & 1.73            &                      & \textbf{4.70}  & \textbf{3.32}   & 2.59            & \textbf{0.58}   & \textbf{2.50}   \\ \hline
\end{tabular}

}
\end{table*}

\subsection{Experimental Setup}
\subsubsection{Metrics} 
We evaluated the performance of our method using a set of task-specific metrics (measured in centimeters), including:
\begin{itemize}[leftmargin=1em]
    \item \emph{Positional Error (PE)}. Measures the mean Euclidean distance in between the predicted joint positions and the ground truth across all frames, indicating the overall reconstruction accuracy.
    \item \emph{End-Effector Positional Error (EEE)}. Computes the average position error, specifically for end-effectors.
    \item \emph{Global Translation Error (GTE)}. Measures the global root translation error between the predicted and ground-truth skeletons.
    \item \emph{Absolute Jitter Error (AJE)}. Captures temporal instability via the absolute difference between predicted motion's mean joint jerk and the evaluation dataset's inherent jitter. Lower AJE indicates more similarity to the ground-truth motion. Measured in $km/s^2$.
    \item \emph{Geometric Structure Error (GSE)}. Measures pose plausibility by comparing the mean pairwise distances between all joints to the ground truth. This metric quantifies how closely the predicted skeleton’s joint relationships match the ground truth. %
    \review{\item \emph{SIP Error}. Quantifies the mean orientation error of the upper arms and legs in the global space. Measured in degrees.}
    \review{\item \emph{Mesh Error}. Measures the average Euclidean distance between the predicted and ground-truth mesh vertices.}
\end{itemize}

\subsubsection{Baselines}
For a more extensive and thorough comparison, we designed the following three baselines: \\[0.1cm]
\textbf{MDS + Procrustes.} The Multidimensional Scaling (MDS) algorithm is a classic technique that projects high-dimensional data into a lower-dimensional space, with the aim of preserving pairwise distances. Procrustes~\citep{Goodall:1991} aims to align two shapes by translating, rotating, and scaling one to best match the other, minimizing the distance between corresponding points. First, we run MDS on the input $\tilde{D}^{[t+1]}$ to approximate $\hat{P}^{[t+1]}$. To align the orientation, we then apply Procrustes between $\hat{P}^{[t]}$ and $\hat{P}^{[t+1]}$. \\[0.1cm]
\textbf{\modelNoSpace-Geo.} Instead of applying generation in the pairwise-distance space, this proposed baseline is applied directly in the 3D geometric space. In practice, a prediction $\hat{P}^{[t]}$ is directly embedded and fed into the Transformer to predict the next pose. \\[0.1cm]
\review{\textbf{\modelNoSpace-feet.} Since prior works for inertial-based sparse mocap often use knees instead of feet~\citep{Yi:2022,Armani:2024:UIP,Rempe:2021}, we train an additional variant to match that exact setting, termed \emph{\modelNoSpace-feet}. However, we argue that our setting is more robust, and specifically benefits more in foot-ground contact accuracy (see~\cref{tab:cf_comparison}).}
\begin{figure}[h]
    \centering
    \includegraphics[width=1\columnwidth]{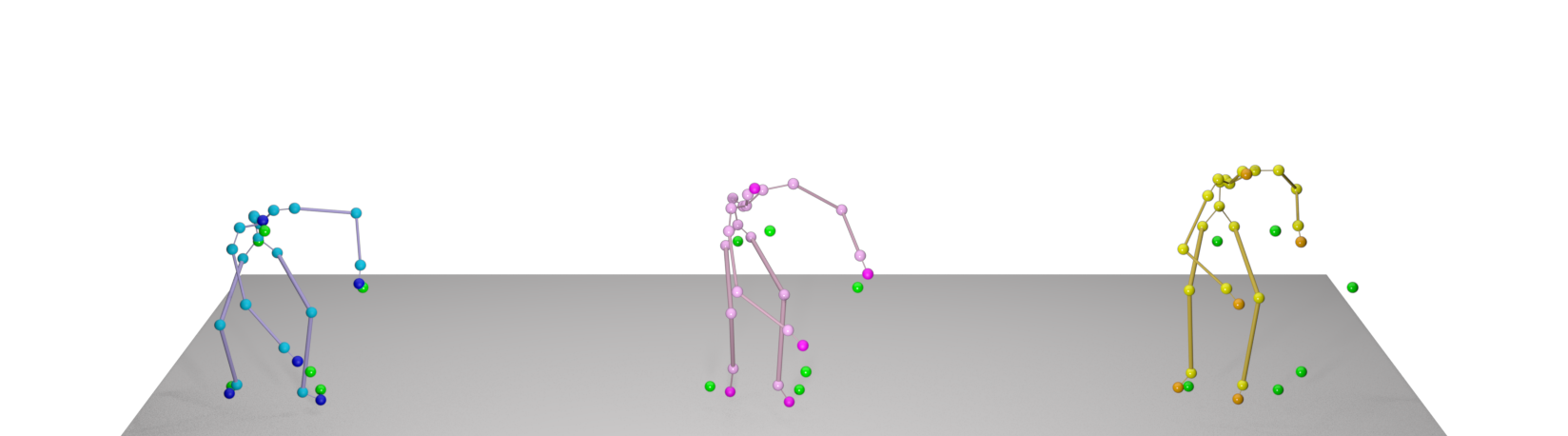}
    \includegraphics[width=1\columnwidth]{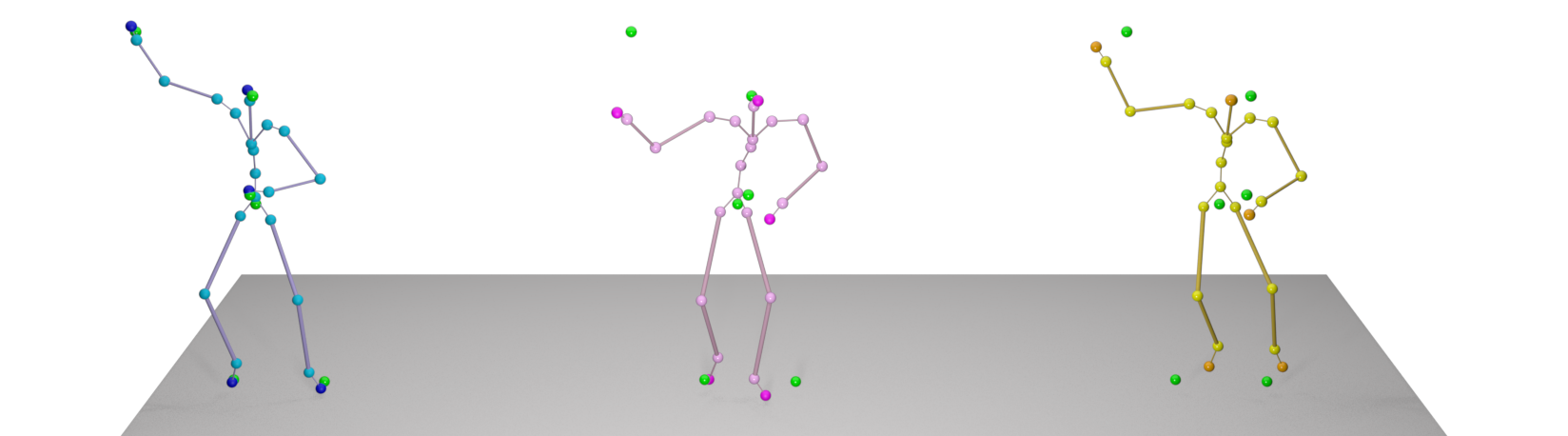}
    \includegraphics[width=1\columnwidth]{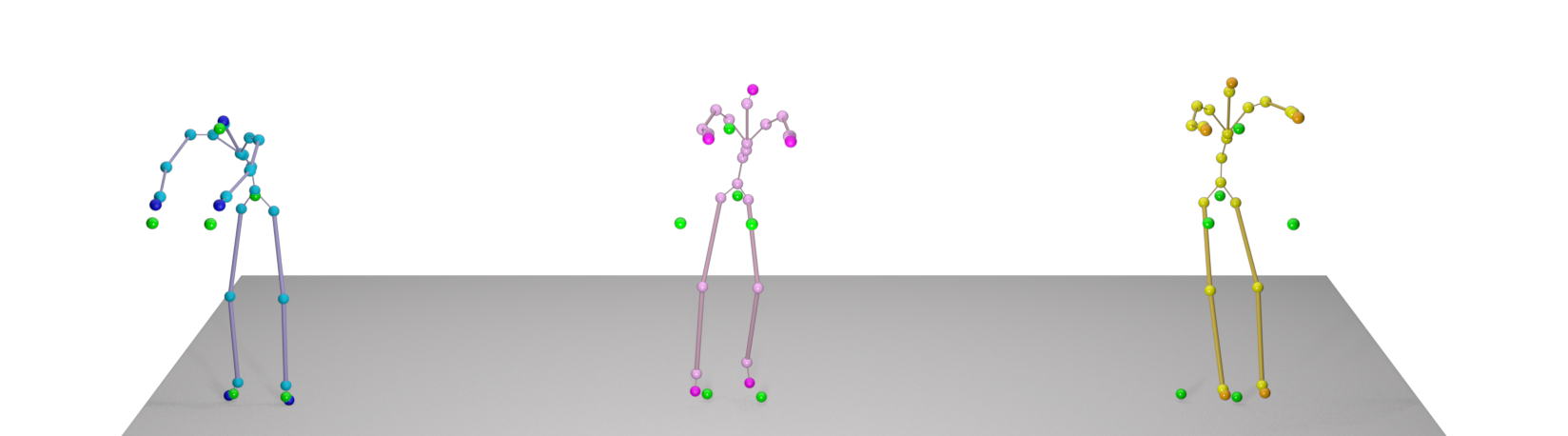}
    \caption{\textbf{Qualitative Comparison of End-Effector Positioning} (from the DanceDB dataset). Note that our method (\textcolor{cyan}{blue}) produces poses where the end-effectors (darker shades) align more closely with the original sensor target positions (green spheres), compared to UIP (\textcolor{magenta}{magenta}) and PIP (\textcolor{yellow}{yellow}).}
    \label{fig:ee_vis_DanceDB}
\end{figure}

\subsection{Comparison with Previous Methods}
\begin{figure}[h]
    \centering
    \includegraphics[width=0.49\columnwidth]{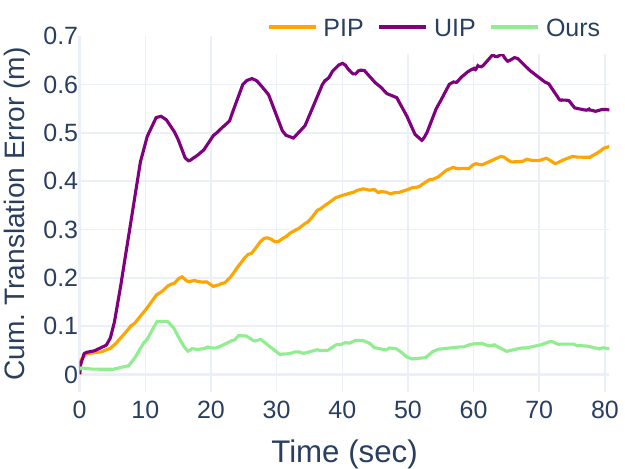}
    \includegraphics[width=0.49\columnwidth]{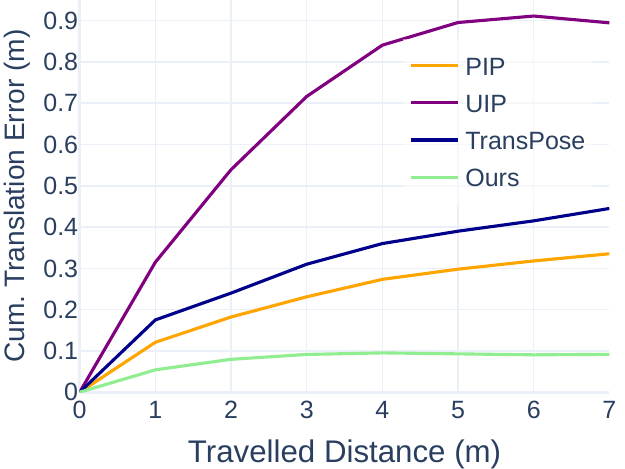}
    \caption{\textbf{Cumulative Drift Error Analysis.} Unlike inertial-based methods, our approach is robust to cumulative error, both in time (left) and distance (right). By utilizing distance measurements and incorporating additional reference sensors as anchors, we can estimate the global position of the subject at every timestamp with minimal reliance on previous predictions.}
    \label{fig:Cumulative Drift}
\end{figure}

\begin{figure}[h]
    \centering
    \includegraphics[width=0.9\columnwidth]{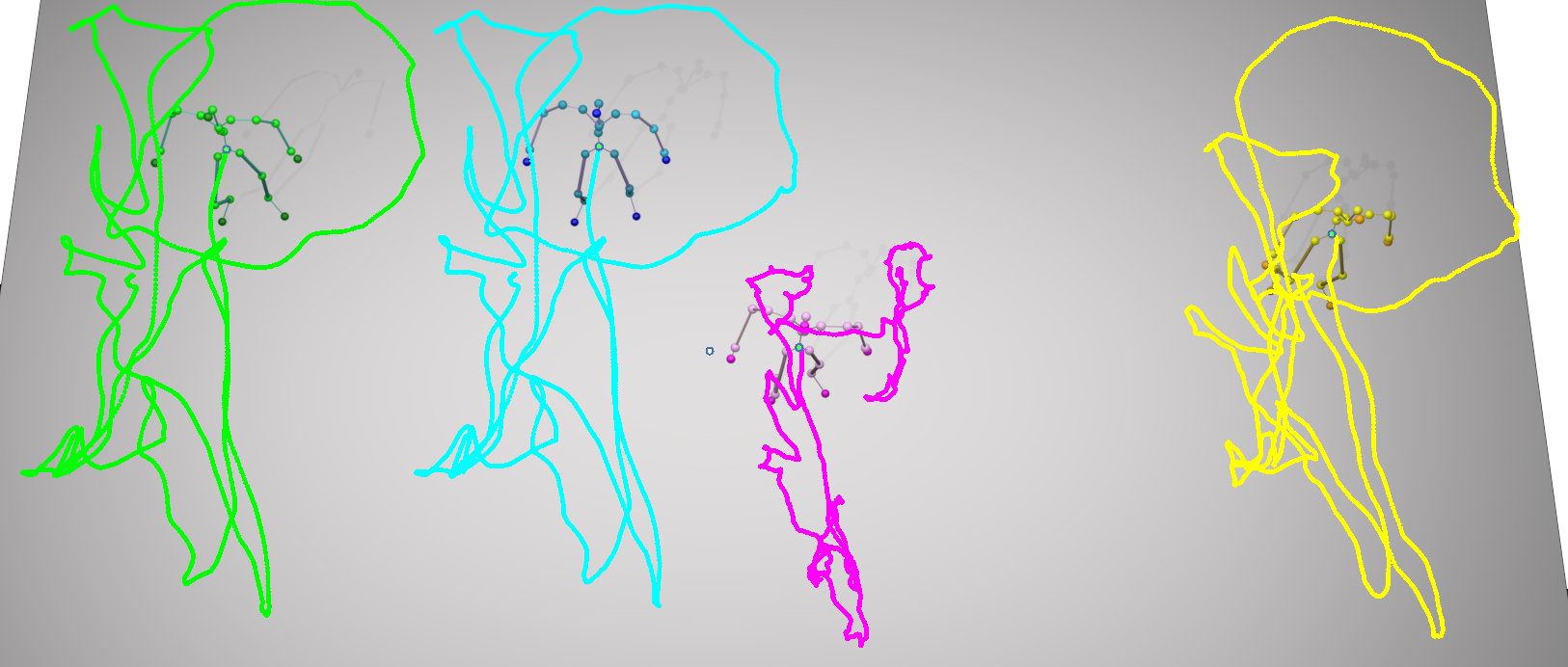}
    \caption{\textbf{Comparison of global root position reconstruction accuracy} (from the TotalCapture dataset). Our method (\textcolor{cyan}{blue}) reconstructs the global root positions closer to the original (\textcolor{green}{green}), compared to UIP (\textcolor{magenta}{magenta}) and PIP (\textcolor{yellow}{yellow}).}
    \label{fig:comp_root_TT}
\end{figure}

We conduct an empirical analysis of \modelNoSpace's performance, comparing it against previous works (UIP~\citep{Armani:2024:UIP}, PIP~\citep{Yi:2022}, TIP~\citep{Jiang:2022}, TransPose~\citep{Yi:2021}, \citep{Liu:2025:UMotion}), and our proposed baselines. In~\cref{tab:comparisons}, we evaluate our method using synthesized PWD derived from three mocap datasets: TotalCapture~\citep{Trumble:BMVC:2017}, DIP-IMU~\citep{Huang:2018}, and DanceDB~\citep{Aristidou:2019:JOCCH}, each excluded from our AMASS~\citep{Mahmood:2019:AMASS} training set. Remarkably, our approach is sufficient to achieve state-of-the-art performance using only PWD, outperforming existing inertial-based methods. Unlike prior work that estimates body poses from individual measurements, our method constructs a decentralized network, enabling the simultaneous reconstruction of both geometric structure and global position at each timestamp. It requires no body-specific calibration or initial pose for evaluation. %
As shown in \cref{fig:ee_vis_DanceDB}, \model significantly improves the position of the end effector, reducing the error by up to $70\%$. This design substantially mitigates cumulative drift, a common limitation of previous methods (\cref{fig:Cumulative Drift}). \model further reduces GTE by $\sim$66\% (\cref{tab:comparisons}), with \cref{fig:comp_root_TT} demonstrating its superior accuracy in global position tracking. 

\begin{table}[t]
\centering
\caption{\textbf{Quantitative Results on Real UWB Data}. We evaluate and compare our model on two benchmark databases: 1) MA-DB, our evaluation dataset and 2) UIP-DB, the test split of the dataset released by UIP. %
For fair comparison on UIP-DB, we report the numbers produced by our \modelNoSpace-knees vriant.}
\label{tab:comparison_Real_UWB}
\vspace{-0.5em}
\resizebox{\columnwidth}{!}
{
\begin{tabular}{lccccccc}
\hline
\multirow{2}{*}{Method} & \multicolumn{3}{c}{MA-DB (feet)}                    & \multicolumn{1}{l}{} & \multicolumn{3}{c}{UIP-DB (knees)}                 \\ \cline{2-4} \cline{6-8} 
                        & \scriptsize EEE & \scriptsize GTE & \scriptsize AJE & \multicolumn{1}{l}{} & \scriptsize PE & \scriptsize GTE & \scriptsize AJE \\ \cline{1-4} \cline{6-8} 
UIP                     & \multicolumn{3}{c}{\multirow{2}{*}{-}}              &                      & 10.55          & 14.19           & 0.24            \\
UMotion                 & \multicolumn{3}{c}{}                                &                      & 10.33          & -               & 0.33            \\
MDS + Proc.             & 60.48           & 57.96           & \textbf{0.21}   &                      & -              & -               & -               \\
WiP-Geo                 & 28.12           & 34.55           & 0.34            &                      & 7.43           & 3.52            & 0.22            \\ \hline
WiP                     & \textbf{22.80}  & \textbf{34.36}  & 0.35            &                      & \textbf{6.88}  & \textbf{6.84}   & \textbf{0.22}   \\ \hline
\end{tabular}
}
\end{table}

\begin{table}[t]
\centering
\caption{\review{\textbf{Rotation- and Mesh-related evaluation on Noisy TotalCapture and UIP-DB}. We use SMPLify3D to estimate joint local rotations from the global joint positions predicted by \modelNoSpace. Consequently, \model attains competitive performance relative to prior methods.}}
\label{tab:rot_comparison}
\vspace{-0.5em}
\resizebox{\columnwidth}{!}
{
\begin{tabular}{lcccccc}
\hline
\multirow{2}{*}{Method} & \multirow{2}{*}{\begin{tabular}[c]{@{}c@{}}Inp. Measur. \\ Type\end{tabular}} & \multicolumn{2}{c}{TotalCapture}                               & \multicolumn{1}{l}{} & \multicolumn{2}{c}{UIP-DB (test)}                              \\ \cline{3-4} \cline{6-7} 
                        &                                                                               & \scriptsize SIP Err & \multicolumn{1}{l}{\scriptsize Mesh Err} &                      & \scriptsize SIP Err & \multicolumn{1}{l}{\scriptsize Mesh Err} \\ \hline
TransPose               & O+A+V                                                                         & 16.58               & -                                        &                      & -                   & -                                        \\
TIP                     & O+A+V                                                                         & 13.22               & 7.42                                     &                      & 33.01               & -                                        \\
PIP                     & O+A+V                                                                         & 15.93               & 6.80                                     &                      & 30.47               & 14.57                                    \\
UIP                     & O+A+V+D                                                                       & 11.32               & 5.09                                     &                      & 23.85               & 14.32                                    \\
UMotion                 & O+A+V+D                                                                       & \textbf{10.76}      & \textbf{4.94}                            &                      & 25.69               & \textbf{11.68}                           \\ \hline
WiP + SMPLify3D         & D                                                                             & 11.63               & 6.53                                     &                      & \textbf{22.37}      & 12.01                                    \\ \hline
\end{tabular}
}
\end{table}

\begin{table}[t]
\centering
\caption{\review{\textbf{Foot-ground contact analysis}. We highlight the potential of our proposed sensor placement by evaluating foot-ground contact accuracy across three known benchmarks.}}
\label{tab:cf_comparison}
\vspace{-0.5em}
\resizebox{\columnwidth}{!}
{
\begin{tabular}{lccccc}
\hline
Method          & TotalCapture  & \multicolumn{1}{l}{} & DIP-IMU       & \multicolumn{1}{l}{} & UIP-DB (test) \\ \hline
UIP (knees)     & 0.79          &                      & 0.73          &                      & 0.81          \\
UMotion (knees) & 0.84          &                      & 0.90          &                      & 0.71          \\
WiP (knees)     & 0.80          &                      & 0.86          &                      & 0.77          \\
WiP (feet)      & \textbf{0.97} &                      & \textbf{0.96} &                      & -             \\ \hline
\end{tabular}
}
\end{table}

\subsection{Evaluation on Raw UWB Measurements}
In~\cref{tab:comparison_Real_UWB}, we present results on real-world UWB data. Our evaluation covers two datasets: MA-DB (our evaluation set) and UIP-DB, released by UIP. \review{To close the gap between the largely different conditions these two datasets were captured, we filter irrelevant metrics (PE in MA-DB, as it was captured using only six sensors, EEE in UIP-DB, as it captured with sensors placed on the knees).}
\review{The table shows that \model outperforms both UIP and our baselines. We find it important to emphasize that the increased error on MA-DB primarily results from minimal preprocessing and noise filtering (unlike UIP-DB, that uses Extended Kalman Filter and other sensor fusion techniques). We further highlight that this was done intentionally to preserve real-time performance, prevent external bias and to evaluate \model directly on raw measurements.}

\review{\subsection{Evaluation on Rotation-/Mesh-based Metrics}
To enable \model to support human skeletal hierarchy representation for character animation and mesh recovery (e.g., the SMPL parametric model), one can employ an off-the-shelf IK solver to estimate joint local rotations and mesh vertex positions. One such technique is SMPLify3D, introduced by \citet{Zuo:2021:Sparsefusion}. SMPLify3D leverages a human motion prior to estimate joint hierarchy rotations consistent with the SMPL model. For rotation- and mesh-based comparisons, please refer to~\cref{tab:rot_comparison}. As shown in the table, \model achieves competitive performance on both TotalCapture and UIP-DB datasets, despite relying solely on PWD data, in contrast to previous methods that utilize both PWD and IMU inputs. It is worth noting that, although our model achieves lower joint position errors, additional inaccuracies are introduced by SMPLify3D through its pose optimization process.} %

\review{\subsection{Lower Body Sensor Placement (Knees vs. Feet)}
Most previous works place sensors, typically IMUs, on the knees. For inertial-based methods, this choice is reasonable, as it effectively captures leg bending. However, we hypothesize that this configuration compromises foot–ground contact performance, an essential aspect of natural motion modeling. To address this, we place the lower-body sensors on the feet instead of the knees. To evaluate the effectiveness of our proposed approach, we measure foot–ground contact accuracy. Following prior work, we define a foot as being in contact with the ground when both the toe and heel joints (i) have a velocity below 0.2 m/s and (ii) are within 10 cm of the ground in the vertical direction. As shown in \cref{tab:cf_comparison}, knee-based methods (UIP, UMotion, and \modelNoSpace-knees) exhibit substantial errors in estimating foot-ground contact, whereas \modelNoSpace-feet returns significantly higher accuracy, achieving the lowest contact error and the most stable and realistic foot–ground interactions.
}

\subsection{Ablation Study}

\review{To evaluate the contribution of each building blocks of \modelNoSpace, we perform an extensive ablation studies on two datasets: (1) TotalCapture with synthetically added noise, used to examine robustness to jitter, and (2) UIP-DB, consisting of real but preprocessed motion data, used to assess performance under realistic conditions. Due to each dataset sensor placement setting, we evaluate \modelNoSpace-feet and \modelNoSpace-knees on TotalCapture and UIP-DB, respectively. As shown in \cref{tab:ablation}, each component contributes uniquely and complementarily. For example, removing the STJ-SA layers reduces jitter but compromises motion flexibility, increasing \review{end-effector} errors. On the other hand, excluding the denoising stage makes the model more `input-guided' rather than motion-guided, resulting in increased jitter. It can also be observed that replacing the Gated Cross-Attention module with standard Cross-Attention layers enforces a constant balance on the PWD measurements, resulting in lower joint position error but increased jitter, leading to shakier and less natural motion. Since UIP-DB is preprocessed, we found it more effective to omit the denoising stage on the \modelNoSpace-knees variant, and train it end-to-end, including STJ-SA layers, on clean data. It is important to note that while \model may not achieve the best performance in every individual metric, it yields the strongest overall results.}

\begin{table}[t]
\centering
\caption{\review{\textbf{Ablation Study on \modelNoSpace's Components.} We report position-related metrics on Noisy TotalCapture and UIP-DB.}}
\label{tab:ablation}
\vspace{-0.5em}
\footnotesize
\def\arraystretch{1.2}
\resizebox{\columnwidth}{!}{
\begin{tabular}{lccccccccc}
\hline
                         & \multicolumn{4}{c}{TotalCapture (+noise)}                                                                                   &  & \multicolumn{4}{c}{UIP-DB (real)}                                                                                                           \\ \cline{2-5} \cline{7-10} 
\multirow{-2}{*}{Method} & \scriptsize PE & \scriptsize EEE              & \scriptsize AJE                      & \scriptsize SGE                      &  & \scriptsize PE              & \scriptsize EEE              & \scriptsize AJE                   & \scriptsize SGE                            \\ \hline
WiP                      & 4.62           & \textbf{6.21}                & 2.56                                 & \textbf{1.86}                        &  & \textbf{6.88}               & \textbf{15.99}              & 0.22                              & \textbf{6.32}                             \\
\quad w/ denoise ft.     & 5.12           & 6.80                         & 1.42                                 & 2.11                                 &  & \multicolumn{4}{c}{-}                                                                                                                       \\ \hline
\quad w/o STJ-SA         & 5.17           & 8.75                         & \textbf{0.71}                        & 2.36                                 &  & 7.42                        & 19.85                        & 0.22                              & 8.16                                       \\
\quad w/o PWD-Aux        & 4.66           & 10.89                        & 2.53                                 & 3.03                                 &  & 8.10                        & 23.81                        & \textbf{0.21}                     & 9.83                                       \\
\quad w/o gating         & \textbf{4.27}  & 9.35                         & 2.34                                 & 2.39                                 &  & 7.14                        & 19.72                        & 0.22                              & 8.71                                       \\ \hline
\end{tabular}
}
\end{table}

\section{Conclusions}

We introduce a novel mocap system that reconstructs full-body 3D poses from sparse pairwise distances, specifically from UWB, without relying on external cameras or pre-calibrated body models. At the core of our approach is a compact Transformer Decoder architecture trained to denoise and interpret temporal PWD sequences for accurate pose estimation. Through extensive experiments, including in-the-wild recordings and tracking of both humans and animals, we demonstrate that our system achieves smooth, high-fidelity reconstructions at real-time frame rates, outperforming existing inertial baselines. Our solution is portable and shape-independent with potential applications in animation and robotics.

\paragraph{Limitations} The Ciholas DWTAG devices currently exhibit a significant level of noise, which affects the overall precision of measurements. However, we anticipate that UWB technology will continue to evolve in the coming years, leading to reductions in signal noise and improvements in hardware stability and performance. These advancements will, in turn, unlock new opportunities for system development and refinement. Our work provides a robust algorithmic foundation designed to capitalize on these technological improvements. As UWB technology advances, we expect the accuracy and reliability of our method to improve correspondingly.

\paragraph{Future Work} To address challenges such as noise sensitivity and NLoS conditions, future work may explore advanced distributed networking strategies. These include organizing sensors into cliques with direct line-of-sight, employing pairwise incremental topologies, or applying multilateration using fixed anchor stations. Another promising direction is to evaluate the system's performance in multi-subject scenarios, where multiple individuals are equipped with UWB sensors, leading to a denser and potentially more informative set of PWD measurements.

\bibliographystyle{ACM-Reference-Format}
\bibliography{Bibliography}

\appendix
\newpage

\section{Appendices}
\label{sec:Appendices}

We extend our analysis of \modelNoSpace. Section~\ref{ssec:wip_denoise} evaluates the pairwise distances (PWD) produced by our model. Section~\ref{ssec:pwd_as_denoiser} examines the robustness of \modelNoSpace's to noisy inputs, and Section~\ref{ssec:stj-sa} provides qualitative results for the newly introduced Spatio-Temporal Joint Self-Attention (STJ-SA) layer.
\begin{figure}[h]
    \centering
    \includegraphics[width=0.49\columnwidth]{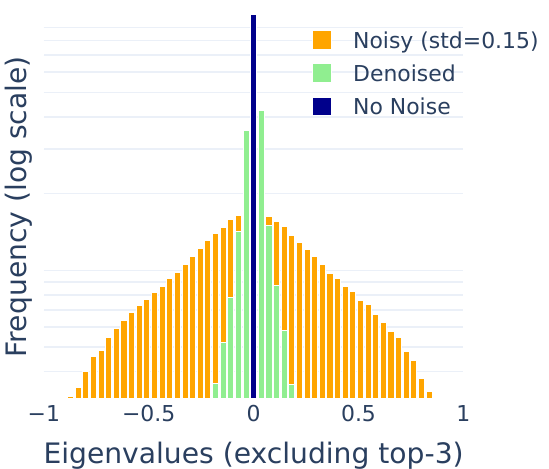}
    \includegraphics[width=0.49\columnwidth]{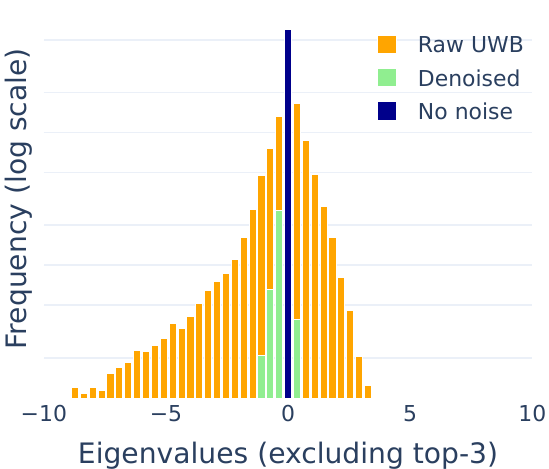}
    \caption{\textbf{Qualitative Analysis of \modelNoSpace's predicted PWDs.} We compute the eigenvalues of the PWDs produced by our PWD auxiliary task head, exclude the top three largest eigenvalues from each matrix, and plot the remaining eigenvalues in a histogram. For completeness and fair comparison we extract only sparse joints (Pelvis, Head, Hands and Feet).}
    \label{tab:eigen_dists}
\end{figure}

\subsection{PWD Denoise Analysis} \label{ssec:wip_denoise}
A well-defined Euclidean PWD matrix from 3D points yields exactly three positive eigenvalues, the rest of which approach zero. Negative eigenvalues indicate non-Euclidean characteristics, such as violations of the triangle inequality, while additional positive eigenvalues suggest the PWD cannot be fully embedded in 3D space. To assess the quality of our prediction of the PWD auxiliary task, we perform an eigenvalue analysis, as shown in~\cref{tab:eigen_dists}. It can be observed that \model produces densely distributed zero-centered eigenvalues, indicating that it implicitly learns the structure of a well-formed Euclidean PWD. \cref{tab:eigen_analysis} reports two metrics that demonstrate the effectiveness of our model predictions: \\[0.1cm]
\emph{Cumulative Explained Variance (CEV) Ratio}: Originally from PCA, the cumulative explained variance ratio is a reliable method to choose the number of components when applying PCA, as it enables dimensionality-reduction while maximizing the retained variance between the original data and its lower-dimensional projection. The CEV is defined as follows:
\begin{equation}
    CEV\left(k\right) = \sum_{i=1}^k\frac{\lambda_i}{\sum_{j=1}^n\lambda_j}
\end{equation}
In a well-defined Euclidean PWD, retaining the largest $k=3$ components is typically sufficient to capture nearly 100\% of the variance in the data, reflecting the underlying three-dimensional structure. We employ this technique to evaluate how much variance is captured by the three largest eigenvalue components. \\[0.1cm]
\emph{Triangle-Inequality Score (TI Score)}: An important property of a well-defined PWD is the satisfaction of of the triangle-inequality for any three indices. We therefore define a new metric for evaluating of this property in an estimated PWD:
\begin{equation}
    \text{TIS}\left(\hat{D}^{[t]}\right) = \frac{1}{J}\sum_{i\ne j\ne k}
    \begin{cases}
        1\hspace{0.5cm} \text{if } \hat{D}^{[t]}_{ij} + \hat{D}^{[t]}_{jk} - \hat{D}^{[t]}_{ik} >= 0\\
        0\hspace{0.5cm} \text{otherwise}
    \end{cases}
\end{equation}
where $\hat{D}^{[t]} \in \mathbb{R}^{J \times J}$ is the estimated PWD for $J$ joints. We compute the $TI$ score for all samples and take the average. A maximum score of $1$ indicates $0\%$ violation.

\begin{table}[ht]
\centering
\caption{\textbf{Qualitative Analysis on PWD Prediction.} \model is trained to estimate well-defined Euclidean PWDs. We define quantitative metrics and compare against different inputs. `GT synth.' refers to perfect PWD computed from optical data, and `Raw input' refers to noisy measurements.} %
\label{tab:eigen_analysis}
\vspace{-0.5em}
\footnotesize
\def\arraystretch{1.2}
\resizebox{\columnwidth}{!}{

\begin{tabular}{lccccc}
\hline
\multirow{2}{*}{Method}        & \multicolumn{2}{c}{TotalCapture noisy}                 & \multicolumn{1}{l}{{\ul }} & \multicolumn{2}{c}{MA-DB}                               \\ \cline{2-3} \cline{5-6} 
                               & CEV Score ($\uparrow$)                & TI Score ($\uparrow$)                   & \multicolumn{1}{l}{}       & CEV Score  ($\uparrow$)                & TI Score  ($\uparrow$)                 \\
\color[HTML]{C0C0C0} GT synth. & \color[HTML]{C0C0C0} 1.00 & \color[HTML]{C0C0C0} 1.000 & \color[HTML]{C0C0C0}       & \color[HTML]{C0C0C0} 1.000 & \color[HTML]{C0C0C0} 1.000 \\
Raw input                      & 0.913                     & 0.909                      &                            & 0.500                      & 0.945                      \\ \hline
\model w/o denoise             & 0.984                     & 0.989                      &                            & 0.990                      & 0.974                      \\
\model                         & \textbf{0.991}            & \textbf{0.993}             &                            & \textbf{0.990}             & \textbf{0.983}             \\ \hline
\end{tabular}
}
\end{table}

\begin{table*}[t]
\centering
\caption{\modelNoSpace's Training Recipes.}
\label{tab:tr_details}
\vspace{-0.5em}
\small
\def\arraystretch{1.2}

\begin{tabular}{lccccccc}
\hline
Training Stage         & \begin{tabular}[c]{@{}c@{}}\#Warmup\\ Steps\end{tabular} & \begin{tabular}[c]{@{}c@{}}\#Training\\ Steps\end{tabular} & Optimizer              & \begin{tabular}[c]{@{}c@{}}Learning\\ Rate\end{tabular} & \begin{tabular}[c]{@{}c@{}}Weight\\ Decay\end{tabular} & \begin{tabular}[c]{@{}c@{}}Effective\\ Batch Size\end{tabular} & \begin{tabular}[c]{@{}c@{}}\#Trainable / Total\\ Parameters (M)\end{tabular} \\ \hline
(I) distance-to-motion & \multirow{2}{*}{$500$}                                   & \multirow{2}{*}{$100K$}                                    & \multirow{2}{*}{AdamW} & \multirow{2}{*}{$1e-4$}                                 & \multirow{2}{*}{$1e-3$}                                & $3072$                                                         & $137.1 /  137.1$                                                             \\
(II) denoising         &                                                          &                                                            &                        &                                                         &                                                        & $256$                                                          & $10.1 / 138.3$                                                               \\ \hline
\end{tabular}
\end{table*}

\subsection{\model as a PWD Denoiser} \label{ssec:pwd_as_denoiser} 
\begin{figure}[t]
    \centering
    \includegraphics[width=.8\columnwidth]{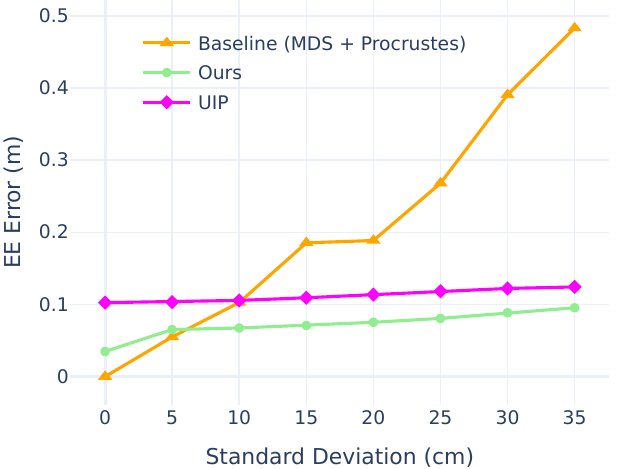}
    \caption{\textbf{End-Effector (EE) Error over Increasing Added Noise.} As input noise gradually increases, EE error also rises. However, \model remains highly robust, consistently keeping EE error low even at high noise levels.}
    \label{fig:noise_level}
\end{figure}

Due to factors such as occlusion, varying distances, and motion speed, the noise in the raw measurements is inherently unpredictable. Thus, robustness to noise and adaptive attention between past predictions and current inputs are essential. In~\cref{fig:noise_level}, we compare the noise tolerance of our model, UIP~\citep{Armani:2024:UIP}, and our classic baseline. Even with noise levels up to 35 cm, our model remains robust and maintains EEE (End-Effector Error) below 10 cm, outperforming both alternatives. Although the baseline excels with perfect distance measurements, it struggles in dense settings, limited by the number of available sensors, and is highly sensitive to noise.

\subsection{STJ-SA Visualization} \label{ssec:stj-sa} 
\begin{figure}[h]
    \centering
    \includegraphics[width=\columnwidth]{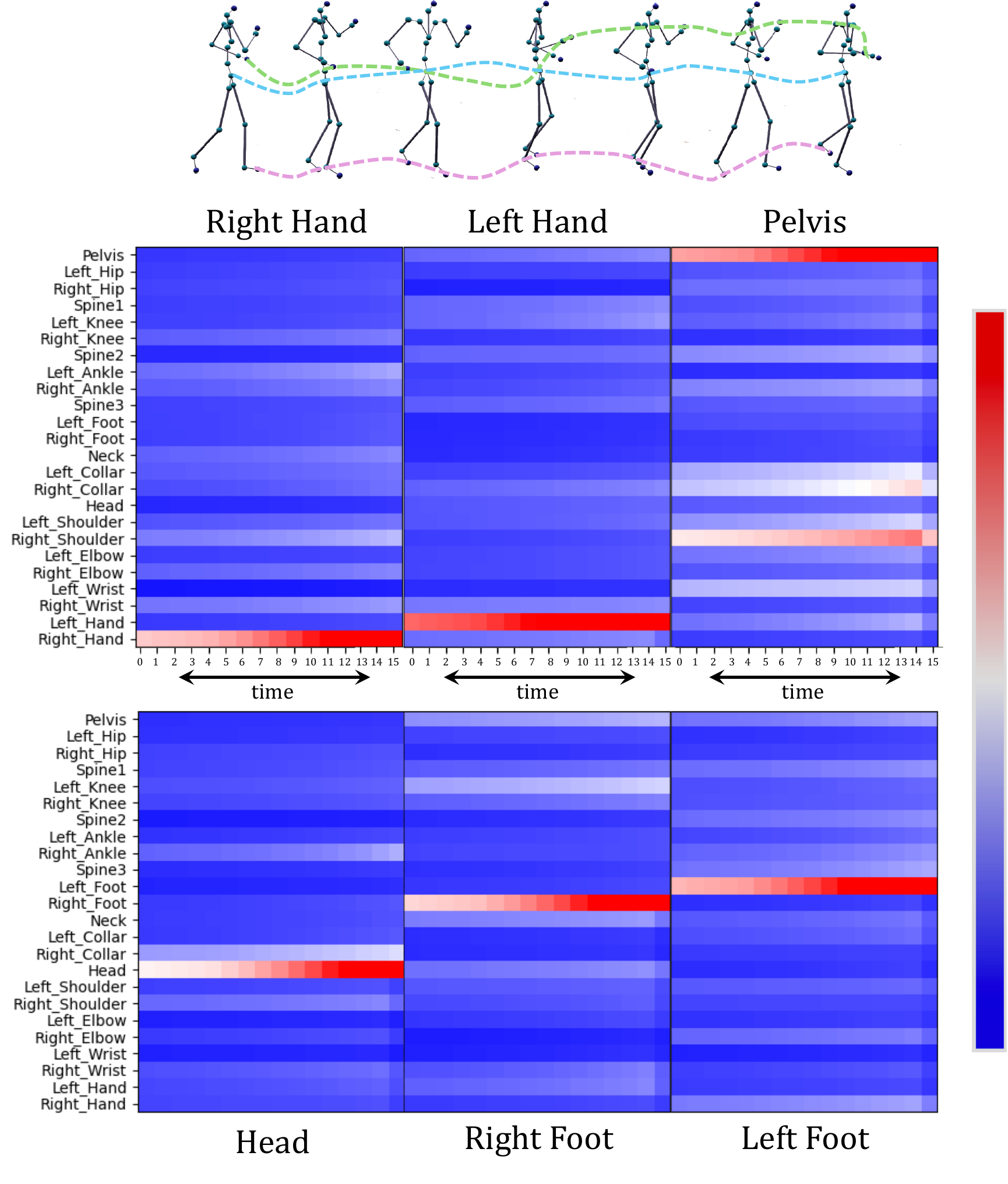}
    \caption{\textbf{STJ-SA Visualization of End-effector Joints.} We extract the attention weight maps from \model's final STJ-SA layer and average them across heads. Red pixels indicate hight attention value (each maps sums up to $1$).}
    \label{fig:attn_vis}
\end{figure}

To further illustrate the effectiveness of our added STJ-SA layers, in~\cref{fig:attn_vis} we perform a qualitative analysis by visualizing the attention maps of key joints over time in the final layer. For each joint, the map indicates how much attention it allocates to other joints across all $T$ timestamps and $J$ joints ($T\cdot J$ in total). The results show a joint $j^{[t]}$ mostly attends its own past instances $j^{[t-i]}$ ($0 \leq i < t$), highlighting the focus on temporal trajectories. Moreover, it can be seen that apart from attention to their own joints, they attend to anatomically-related joints (Pelvis $\rightarrow$ R/L Collars, Right Hand $\rightarrow$ Right Shoulder, Head $\rightarrow$ Right Collar).

\section{Implementation Details}
\label{sec:Further Implementation Details}

\subsection{PWD Embedding} 
\begin{figure}[h]
    \centering
    \includegraphics[width=.8\columnwidth]{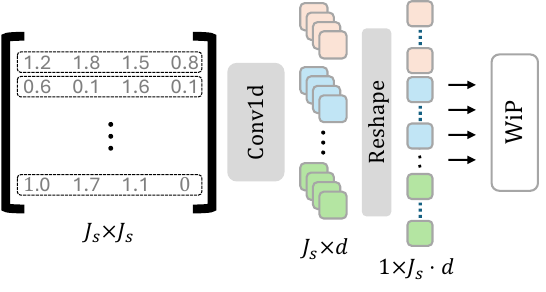}
    \caption{\textbf{PWD Embedding Process.} We preserve the spatial structure in the embedding by applying a 1D convolution with both the kernel size and stride set to $J$.}
    \label{fig:pwd_embed}
\end{figure}

Encoding pairwise distances (PWDs) as input to a Transformer model is non-trivial. To this end, we propose a novel PWD embedding strategy. In \modelNoSpace, each token represents a spatially structured embedding for the entire skeleton. To preserve joint-specific spatial relationships, we derive the embedding for joint $j$ from the corresponding row $\tilde{D}_j$ of the PWD matrix. This is implemented using a row-wise 1D convolution layer with a kernel and stride sizes equal to $J$ and output channels $d$. This operation transforms each joint into a $d$-dimensional embedding. The resulting activations are then reshaped into a spatially coherent pose embedding $\text{Emb}\left(\tilde{D}^{[t]}\right) \in \mathbb{R}^{J\times d}$, encodes both global skeletal context and joint-specific spatial relationships (see~\cref{fig:pwd_embed} for illustration).

\subsection{Training} In both training stages, we utilize the AMASS motion collection~\citep{Mahmood:2019:AMASS}, excluding TotalCapture \citep{Trumble:BMVC:2017} for evaluation. When evaluating on DanceDB \citep{Aristidou:2019:JOCCH}, we remove it from the training set and include TotalCapture.
Following the literature, we train our model using a teacher-forcing strategy to ensure stability and improve convergence. More training details are shown in~\cref{tab:tr_details}. \review{For loss component weights $\lambda_i$, we set: $\lambda_{dd}=1.0,\lambda_{pd}=1.0,\lambda_{refs}=0.5,\lambda_{cons}=0.5,\lambda_{velo}=0.1,\lambda_{rigidity}=1.0,\mathcal{R}_{gravity}=0.05$.}

\subsection{Evaluation}
For the generation process, to predict the next pose at timestamp $t+1$, we calculate $\hat{D}^{[t]} \in \mathbb{R}^{J \times 3}$ by calculating the pairwise distances of the last predicted pose $\hat{P}^{[t]} \in \mathbb{R}^{J \times J}$. We then concatenate this with the predicted sequence from the previous pass $\hat{D}^{[t-w]}, \ldots, \hat{D}^{[t-1]}$ for the next pass. To further reduce jitter, we found it effective to apply a weighted average over the past $\left\lfloor \frac{w}{2} \right\rfloor$ poses, using a half-normal Gaussian kernel with a low standard deviation. For a fair comparison, we apply this in all of our experiments and report the best-performing option.

\section{Hardware Details}
\label{sec:Hardware Details}

We use DWTAG100 sensors~\citep{CiholasDWTAG} from Ciholas Inc., which support high-frequency PWD measurements and meet the baseline requirements for articulated motion capture at 30Hz. Localization uses TDoA and the Ciholas Data Protocol for low-latency communication. DWTAG100 tags communicate using the Ciholas Data Protocol, a low-latency binary protocol that transmits real-time timing and sensor data, while localization is computed via TDoA measurements. The sensors are lightweight (17g), battery-powered UWB transceiver, with compact dimensions (37.2 mm diameter, 13.4 mm height) and include onboard IMU sensors for inertial motion capture. Global 3D localization can be achieved through communication with fixed DWETH111 anchors~\citep{CiholasDWETH}, allowing high-precision tracking in unstructured environments.
\begin{figure}[t] 
    \centering 
    \includegraphics[width=0.75\columnwidth]{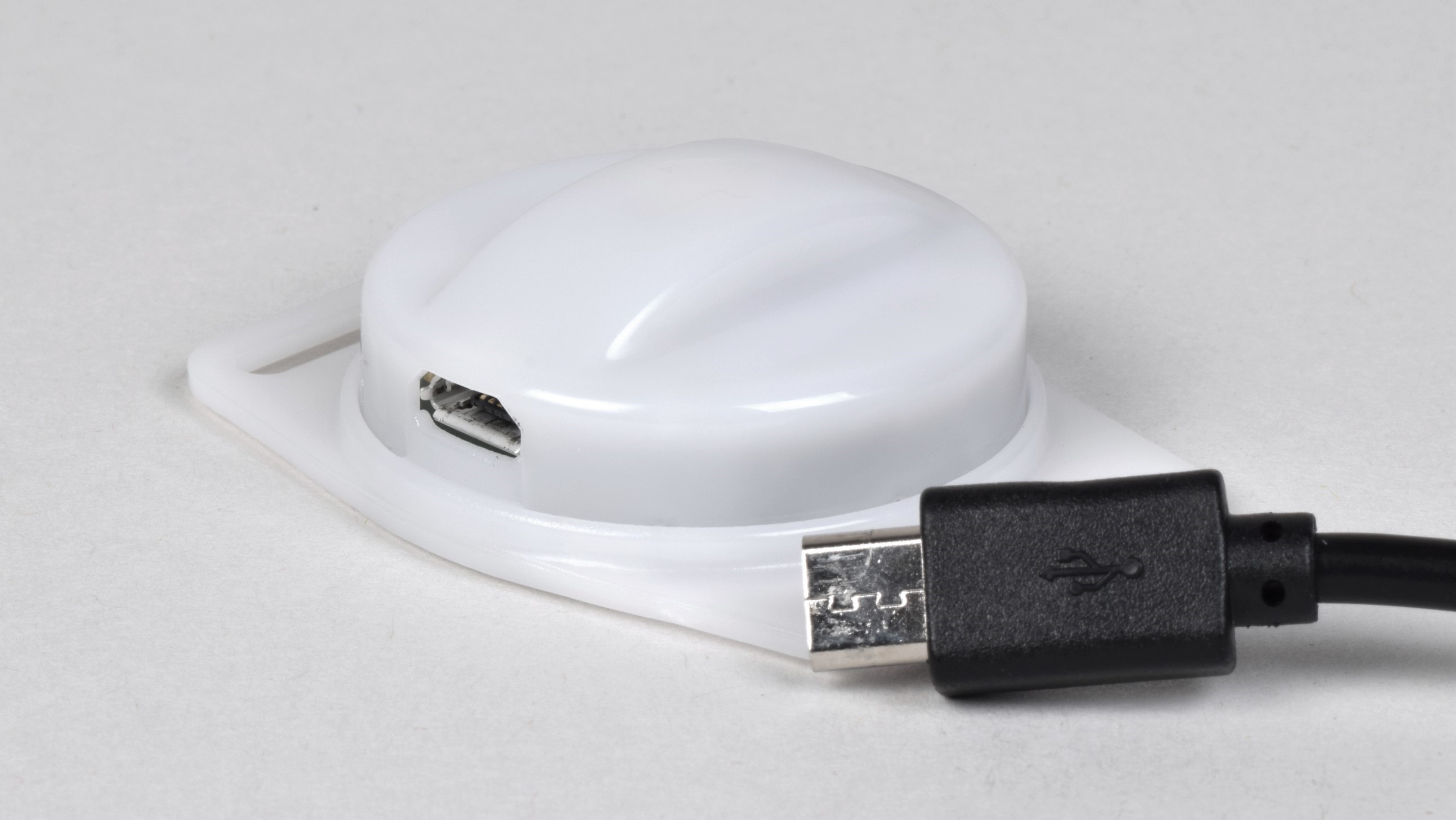} 
    \caption{Ciholas DWTAG100 sensors.} 
    \label{fig:DWTAG100} 
\end{figure}

\end{document}